\documentclass[10pt,twocolumn,letterpaper]{article}
\usepackage[final]{cvpr}
\usepackage{3dv24}

\definecolor{uclablue}{rgb}{0.15, 0.45, 0.68}
\definecolor{cvprblue}{rgb}{0.21,0.49,0.74}
\definecolor{stageone}{rgb}{0.47, 0.63, 0.68}
\definecolor{stagetwo}{rgb}{1, 0.56, 0.31}
\definecolor{colorm}{rgb}{0.47, 0.30, 0.15}
\definecolor{depthm}{rgb}{0.28, 0.57, 0.54}
\definecolor{normalm}{rgb}{0.88, 0.50, 0.71}

\acrodef{sdf}[SDF]{signed distance function}
\acrodef{nerf}[NeRF]{neural radiance fields}
\acrodef{cd}[CD]{Chamfer Distance}
\acrodef{nc}[NC]{Normal Consistency}
\acrodef{rep}[SSR]{Single-view Implicit Surfaces and Radiance Fields}
\acrodef{roi}[ROI]{region of interest}

\newcommand{\modelname}{{\texttt{SSR}}\xspace}
\newcommand{\threedfront}{\mbox{3D-FRONT}\xspace}
\newcommand{\threedfuture}{\mbox{3D-FUTURE}\xspace}
\newcommand{\sunrgbd}{\mbox{SUNRGB-D}\xspace}
\newcommand{\pixthreed}{\mbox{Pix3D}\xspace}
\newcommand{\shapee}{\mbox{Shap$\cdot$E}\xspace}

\newcommand{\fscore}{\mbox{F-Score}\xspace}
\newcommand{\normalconsist}{Normal Consistency\xspace}
\newcommand{\gt}{\mbox{ground-truth}\xspace}
\newcommand{\sota}{\mbox{state-of-the-art}\xspace}
\newcommand{\stageone}{{\color{stageone}Stage One}\xspace}
\newcommand{\stagetwo}{{\color{stagetwo}Stage Two}\xspace}

\newcommand{\supmat}{\textbf{Sup. Mat.}\xspace}

\def\paperID{228} 
\def\confName{3DV\xspace}
\def\confYear{2024\xspace}

\title{Single-view 3D Scene Reconstruction with High-fidelity Shape and Texture}

\begin{document}
\author{
Yixin Chen$^{1\,\star{}}$, Junfeng Ni$^{2\,\star{}\,\dagger}$, Nan Jiang$^{3\,\dagger}$, Yaowei Zhang$^{1}$, Yixin Zhu$^{3}$, Siyuan Huang$^{1}$
\vspace{3pt}\\
\begin{tabular}{c}\small$^\star{}$Equal contributors\quad{}
\small$^\dagger$ Work done during an internship at BIGAI\\
\small$^1$ National Key Laboratory of General Artificial Intelligence, BIGAI\quad{}
\small$^2$ Tsinghua University\quad{}
\small$^3$ Peking University\vspace{3pt}\\
\url{https://dali-jack.github.io/SSR}
\end{tabular}
}

\twocolumn[{
    \renewcommand\twocolumn[1][]{#1}
    \maketitle
    \centering
    \captionsetup{type=figure}
    \vspace{-18pt}
    \includegraphics[width=\linewidth]{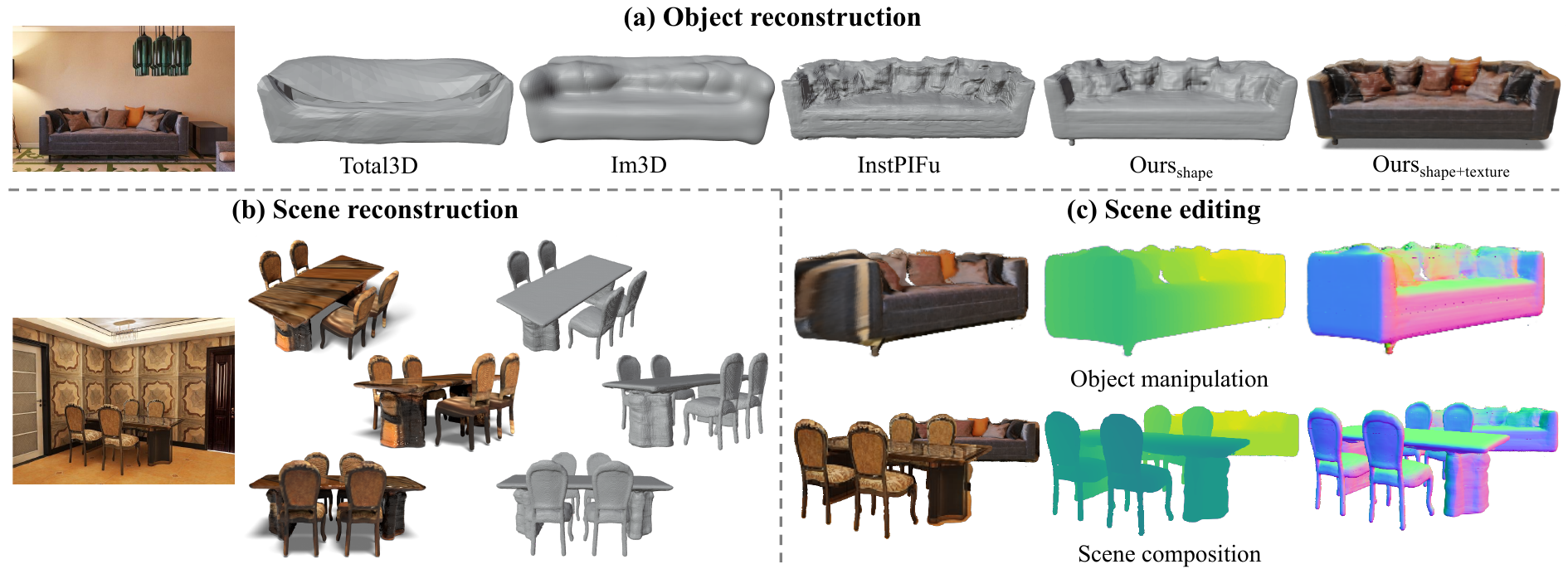}
    \captionof{figure}{\textbf{Single-view 3D scene reconstruction with high-fidelity shape and texture.} (a) Object-level and (b) scene-level reconstruction. Rendering of \textbf{\color{colorm}color}, \textbf{\color{depthm}depth}, and \textbf{\color{normalm}normal} images from the original and novel viewpoints enables 3D scene editing. (c) Object manipulation by rotating the object in (a) and scene composition of (a) and (b).}
    \label{fig:teaser}\vspace{12pt}
}]

\begin{abstract}\vspace{-9pt}
Reconstructing detailed 3D scenes from single-view images remains a challenging task due to limitations in existing approaches, which primarily focus on geometric shape recovery, overlooking object appearances and fine shape details. To address these challenges, we propose a novel framework for simultaneous high-fidelity recovery of object shapes and textures from single-view images. Our approach utilizes the proposed \underline{S}ingle-view neural implicit \underline{S}hape and \underline{R}adiance field (\modelname) representations to leverage both explicit 3D shape supervision and volume rendering of color, depth, and surface normal images. To overcome shape-appearance ambiguity under partial observations, we introduce a two-stage learning curriculum incorporating both 3D and 2D supervisions.
A distinctive feature of our framework is its ability to generate fine-grained textured meshes while seamlessly integrating rendering capabilities into the single-view 3D reconstruction model. This integration enables not only improved textured 3D object reconstruction by $27.7\%$ and $11.6\%$ on the \textit{\threedfront} and \textit{\pixthreed} datasets, respectively, but also supports the rendering of images from novel viewpoints.
Beyond individual objects, our approach facilitates composing object-level representations into flexible scene representations, thereby enabling applications such as holistic scene understanding and 3D scene editing. We conduct extensive experiments to demonstrate the effectiveness of our method.
\end{abstract}

\section{Introduction}\label{sec:intro}

Single-view 3D reconstruction is a challenging task in computer vision that aims to recover a scene's 3D geometry and appearance from a single monocular image. This task holds immense importance as it allows machines to understand and interact with the real 3D world, enabling various applications in virtual reality, augmented reality, and robotics.

The primary obstacle in single-view reconstruction lies in the inherent uncertainties and ambiguities resulting from the limited observations provided by a single image. A model must be able to infer the 3D object shape accurately based on visible regions while also generating a plausible representation of unseen object parts present in the image.

Over the years, various representations and methods have been proposed to tackle this challenge. Early methods in this field utilize 3D bounding boxes to parameterize 3D objects and estimate their size, rotation, and translation~\cite{hedau2009recovering,dasgupta2016delay,huang2018cooperative,du2018learning,chen2019holistic++}. Recent advancements have focused on recovering detailed object shapes using either explicit~\cite{nie2020total3d,gkioxari2019mesh} or implicit~\cite{zhang2021holistic,liu2022towards} representations. However, these approaches suffer from two notable drawbacks. First, they neglect the importance of object textures, which contain essential geometric and semantic details for embodied tasks~\cite{gadre2022cow,huang23vlmaps,ma2023sqa3d} and 3D vision-language reasoning~\cite{chen2020scanrefer,achli2020referit3d,zhao20223dgqa,azuma2022scanqa,chen2021scan2cap}. Second, they often rely solely on image inputs for feature extraction without taking direct textural supervision from them~\cite{nie2020total3d,zhang2021holistic}. Consequently, these models tend to focus insufficiently on geometric subtleties and may learn mean shapes for each object category, leading to challenges in generating smooth and instance-specific details, even when instance and pixel-aligned features are utilized for implicit representation learning~\cite{liu2022towards}.

To address the aforementioned limitations and improve single-view 3D reconstruction, we propose a novel framework that \textbf{simultaneously recovers shapes and textures} from single-view images. Our framework leverages the \underline{S}ingle-view neural implicit \underline{S}hape and \underline{R}adiance field (\modelname) representations. Conditioned on the input image, we extract pixel-aligned and instance-aligned features to predict the \ac{sdf} value using an implicit network and the color value using a rendering network for each query 3D point. By expressing volume density as a function of the \ac{sdf}, our model can be trained end-to-end with \textbf{both 3D shape supervision and volume rendering} of color, depth, and surface normal images.

However, due to shape-appearance ambiguity, simply incorporating rendering supervision may lead to generating realistic textured images but inconsistent underlying geometry~\cite{newcombe2011dtam,colomina2014unmanned}, especially under partial observations. To tackle this issue and achieve improved coordination between 2D and 3D supervision, we propose a carefully designed two-stage learning curriculum. This curriculum balances the rendering and reconstruction losses, allowing our framework to learn a 3D object prior that reconstructs unseen parts from partial observations while capturing pixel-level fine-grained details from the images.

We extensively evaluate our proposed model for single-view object reconstruction on both synthetic \threedfront dataset~\cite{fu20213d} and real \pixthreed dataset~\cite{sun2018pix3d}. The experimental results demonstrate that our method excels in recovering high-fidelity object shapes and textures, significantly outperforming state-of-the-art methods by \textbf{27.7\%} and \textbf{11.6\%} on \threedfront and \pixthreed, respectively. Through thorough ablation studies, we demonstrate the benefits of introducing textural supervision and the importance of the learning curriculum. We show that our model is capable of rendering images from novel viewpoints given single-view inputs, and the quality of the rendered depth and normal is comparable with existing depth and normal estimators~\cite{eftekhar2021omnidata,zamir2020robust}. Finally, we showcase our model's capability in holistic scene understanding and 3D scene editing, allowing for object translation, rotation, and composition of objects in 3D space.

In summary, our work represents a significant advancement in single-view 3D reconstruction, and our contributions are three-fold:
\begin{enumerate}[leftmargin=*]
    \item We propose a novel framework that simultaneously recovers high-fidelity object shapes and textures from single-view images. Our framework leverages the strengths of neural implicit surfaces in shape learning and radiance fields in texture modeling, and seamlessly introduces rendering capabilities into a single-view 3D reconstruction model.
    \item To effectively employ supervision from both 3D shapes and volume rendering, we conduct a thorough analysis and propose a carefully designed two-stage learning curriculum that improves 2D-3D supervision coordination and addresses shape-appearance ambiguity.
    \item Extensive experiments and ablations demonstrate that our proposed method significantly enhances the details of textured 3D object reconstruction, outperforming all state-of-the-art methods. We demonstrate its ability to render color, depth, and normal images from novel viewpoints and its potential to facilitate applications such as holistic scene understanding and 3D scene editing.
\end{enumerate}

\section{Related work}

\paragraph{3D reconstruction from a single Image}

Reconstructing 3D shapes from single images remains a challenging task in indoor scene understanding, and it has spurred the development of relevant datasets~\cite{fu20213d,song2015sun,dai2017scannet,sun2018pix3d} and models~\cite{hoiem2005automatic,dasgupta2016delay,lee2009geometric,han2005bottom}. Early approaches utilized 3D bounding boxes~\cite{hedau2009recovering,huang2018cooperative,du2018learning,huang2019perspectivenet,chen2019holistic++} or retrieved CAD models~\cite{huang2018holistic,mitra2018seethrough,izadinia2017im2cad} to represent objects, but they lacked instance-specific 3D object geometries. Recent methods explored explicit~\cite{nie2020total3d,gkioxari2019mesh} or implicit~\cite{zhang2021holistic,liu2022towards} representations to address these limitations. However, they overlooked object textures, a crucial aspect for semantic-demanding tasks that require pixel-level details. This is often addressed through generative approaches given the 3D shapes~\cite{siddiqui2022texturify,bokhovkin2023mesh2tex}. In this paper, we propose a novel approach that simultaneously recovers detailed 3D geometry and object textures using neural implicit shape and radiance field representation.

Generative methods have also been proposed for single-view 3D reconstruction, utilizing priors learned from large-scale datasets. 2D prior-guided models~\cite{tang2023make,liu2023zero1to3,shen2023anything3d,melas2023realfusion} generate images from novel views and reconstruct objects under a multi-view setting, while the 3D counterpart employs millions of 3D-text pairs to train a conditional generative model~\cite{jun2023shape}. In comparison, our approach leverages benefits from both 3D and 2D supervision in a discriminative way and demonstrates superior capture of high-fidelity 3D structures and details by explicitly modeling the object shapes and textures together.

\paragraph{Neural implicit surfaces representation}

Implicit representations model 3D geometry with neural networks in a parametrized manner~\cite{park2019deepsdf,niemeyer2019occupancy,eslami2018neural,liu2020neural}. Unlike explicit representations (such as point cloud~\cite{qi2017pointnet,achlioptas2018learning}, mesh~\cite{gkioxari2019mesh,pfaff2020learning}, voxels~\cite{wu2016learning,kar2017learning}), implicit representations are continuous, high spatial resolution, and have constant memory usage. However, most existing work~\cite{chen2019learning,nie2020total3d,mescheder2019occupancy,niemeyer2020differentiable,park2019deepsdf} conditions neural implicit representation on global image features, which improves memory efficiency but compromises on preserving details, leading to retrieval-like results. Even when instance and pixel-aligned features are utilized for implicit representation learning from a single view~\cite{saito2019pifu,liu2022towards,xu2019disn}, the model may fail to capture higher-order relationships between 3D points and lack incentives to capture geometric details reflecting pixel-level image details. In this paper, we address this limitation by employing neural implicit shape and radiance field representation, which benefits from both 3D shape supervision and volume rendering, allowing the model to learn geometric and appearance details jointly.

\paragraph{Surface representation learning}

Recent advances in implicit volume rendering (\eg, \ac{nerf}~\cite{martin2021nerf,tewari2020state,kato2020differentiable}) have spurred new efforts in surface representation learning. However, extracting high-fidelity surfaces from learned radiance fields is challenging due to insufficient constraints on the level sets in density-based scene representation. To overcome this limitation, recent methods have combined the benefits of implicit surface and volume rendering-based methods by converting the \ac{sdf} to density and applying volume rendering to train this representation with robustness~\cite{wang2021neus,oechsle2021unisurf,yariv2021volume}. Nevertheless, rendering-based approaches often yield unsatisfactory results in 3D geometry recovery, especially when provided with sparse input views, such as in cluttered indoor scenes. Such failure is rooted in the shape-appearance ambiguity with photo-realistic losses, where an infinite number of photo-consistent explanations exist for the same input image. In this work, we propose an approach that leverages both 3D shape and volume rendering supervision for single-view reconstruction. Moreover, we make the first attempt to investigate how to coordinate these two sources of supervision effectively. To this end, we introduce a two-stage learning curriculum with an incremental increase for the rendering loss weights to achieve improved coordination between the 3D and 2D supervisions and better capture geometric details for textured 3D object reconstruction.

\section{Method}

Given a single image of an indoor scene, our objective is to simultaneously reconstruct the 3D geometry and appearance of all objects present. We build upon existing methods~\cite{liu2022towards,nie2020total3d,zhang2021holistic} for 3D object detection and camera pose estimation, focusing on the reconstruction of fine-grained textured meshes.

\begin{figure*}[ht!]
	\centering
	\includegraphics[width=\linewidth]{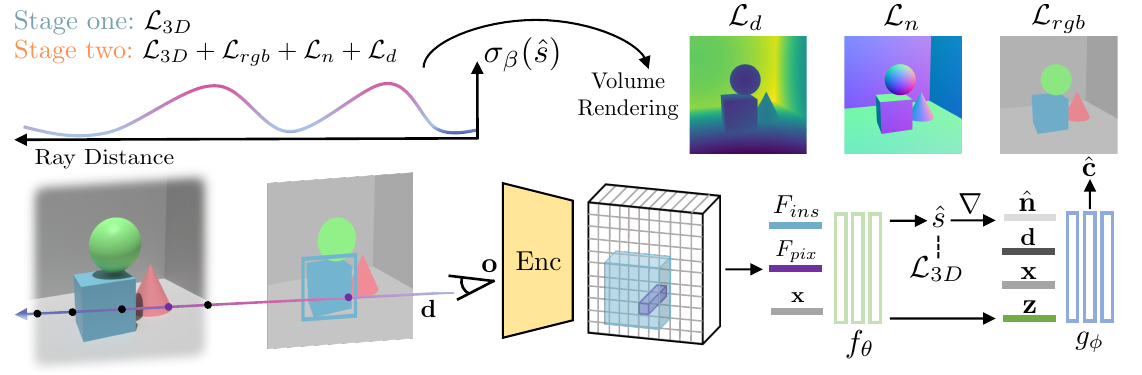}
	\caption{\textbf{Framework overview.} Our framework jointly recovers 3D object shapes and textures from single-view images. Given a query point $\textbf{x}$ along a camera ray with direction $\textbf{d}$, we extract pixel-aligned and instance-aligned features using an image encoder $\mathrm{Enc}$. The implicit network $f_\theta$ predicts the geometry feature $\hat{\textbf{z}}$ and \ac{sdf} value $\hat{s}$, which is then transformed to volume density $\sigma$. The rendering network $g_\theta$ takes the normal $\hat{\textbf{n}}$ and viewing direction $\textbf{d}$ to predict the color value $\hat{\textbf{c}}$. Our learning curriculum consists of two stages: \stageone, which only employs explicit \ac{sdf} supervision, and \stagetwo, where volume rendering supervision is incrementally added.}
	\label{fig:framework}
\end{figure*}

\subsection{Background}\label{sec:background}

\paragraph{Neural implicit surfaces with \ac{sdf}}

We utilize neural implicit surfaces with \ac{sdf} to represent 3D geometry. The \ac{sdf} provides a continuous function that yields the distance $s$ to the closest surface for a given point $\textbf{x}$, with the sign indicating whether the point lies inside (negative) or outside (positive) the surface:
\begin{equation}
    \text{SDF}(\textbf{x})\ =\ s : \textbf{x} \in \mathbb{R}^3,\ s \in \mathbb{R}.
\end{equation}
The zero-level set of the \ac{sdf} function $\Omega = \{\textbf{x} \in \mathbb{R}^3 \mid \ac{sdf}(\textbf{x})=0\}$ implicitly represents the surface.

\paragraph{Volume rendering of implicit surfaces}

To enable optimization with differentiable volume rendering, we convert the neural implicit surface representation \ac{sdf} to density \(\sigma\)~\cite{wang2021neus,oechsle2021unisurf,yariv2021volume}. The conversion is performed using a learnable parameter \(\beta\) as follows:
\begin{equation}
    \sigma_\beta(s) = \begin{cases} \frac{1}{2\beta}\exp(\frac{s}{\beta}) & s \leq 0 \\ \frac{1}{\beta}(1 - \exp(-\frac{s}{\beta})) & s \ge 0 \end{cases}.
\end{equation}

Following the concept of \ac{nerf}~\cite{mildenhall2021nerf}, we sample \(M\) points on the ray \(\textbf{r}\) from the camera center \(\textbf{o}\) to the pixel along the viewing direction \(\textbf{d}\):
\begin{equation}
    \textbf{x}_{\textbf{r}}^i = \textbf{o} + t_{\textbf{r}}^i \textbf{d}, \quad i=1, \ldots, M,
\end{equation}
where \(t_\textbf{r}^i\) is the distance from the sample point to the camera center. We predict the \ac{sdf} value \(\hat{s}\) and color value \(\hat{\textbf{c}}_{\textbf{r}}^{i}\) for each sample point on the ray.

The predicted color \(\hat{C}(\textbf{r})\) for the ray \(\textbf{r}\) can be computed using transmittance \(T_{\textbf{r}}^{i}\) and alpha values \(\alpha_{\textbf{r}}^{i}\):
\begin{equation}
    \hat{C}(\textbf{r}) = \sum_{i=1}^{M} T_{\textbf{r}}^{i}\alpha_{\textbf{r}}^{i}\hat{\textbf{c}}_{\textbf{r}}^{i},
\end{equation}
where \(T_{\textbf{r}}^{i} = \prod_{j=1}^{i-1}(1-\alpha_{\textbf{r}}^{j})\). The alpha value is calculated as \(\alpha_{\textbf{r}}^{i} = 1-\exp(-\sigma_{\textbf{r}}^{i}\delta_{\textbf{r}}^{i})\), and \(\delta_{\textbf{r}}^{i}\) represents the distance between neighboring sample points along the ray. Additionally, we can compute the depth \(\hat{D}(\textbf{r})\) and normal \(\hat{N}(\textbf{r})\) of the surface intersecting the current ray \(\textbf{r}\) as:
\begin{equation}
    \hat{D}(\textbf{r}) = \sum_{i=1}^{M} T_{\textbf{r}}^{i}\alpha_{\textbf{r}}^{i}t_{\textbf{r}}^{i}, \quad  \hat{N}(\textbf{r}) = \sum_{i=1}^{M} T_{\textbf{r}}^{i}\alpha_{\textbf{r}}^{i}\textbf{n}_{\textbf{r}}^{i}.
\end{equation}

\subsection{3D object reconstruction}

Given the input image $I$ of the scene, we aim to recover the 3D shape of the object $\mathcal{O}$, identified by its 2D bounding box, 3D bounding box, and category class.

\paragraph{Feature extraction}

We extract image features $F = \mathrm{Enc}(I)$ using a CNN-based encoder $\mathrm{Enc}$ and utilize both instance-aligned feature $F_{ins}$ and pixel-aligned feature $F_{pix}$ for recovering detailed shapes and textures. $F_{ins}(\mathcal{O})$ is obtained by cropping out the region-of-interest (ROI) features from $F$ based on the 2D bounding box of object $\mathcal{O}$ following He \etal~\cite{he2017mask} and Liu \etal~\cite{liu2022towards}. To obtain the pixel-aligned feature $F_{pix}(\textbf{x})$ for a 3D point $\textbf{x}$, we project $\textbf{x}$ onto the image plane to obtain the corresponding image coordinates $\pi(\textbf{x})$ using the camera intrinsics. The pixel-aligned feature is then obtained through linear interpolation on the feature map, \ie, $F_{pix}(\textbf{x}) = \mathrm{Interp}(F(\pi(\textbf{x})))$.

\paragraph{Implicit and rendering networks}

We parameterize the \ac{sdf} function with an implicit network $f_\theta$, which is a single MLP~\cite{park2019deepsdf,yu2021pixelnerf} taking the instance-aligned feature, pixel-aligned feature, and the position $\textbf{x}$ as input to predict the \ac{sdf} value $\hat{s}$:
\begin{equation}
    \hat{s} = f_\theta(\gamma(\textbf{x}), F_{ins}(\mathcal{O}), F_{pix}(\textbf{x})).
\end{equation}
Here, $\gamma(\cdot)$ represents a positional encoding with 6 exponentially increasing frequencies. The rendering network $g_\phi$ predicts RGB color values $\hat{\textbf{c}}$ for each 3D point using the 3D point $\textbf{x}$, normal $\hat{\textbf{n}}$, viewing direction $\textbf{d}$, and a global geometry feature $\hat{\textbf{z}}$ as input, following Yariv \etal~\cite{yariv2020multiview}:
\begin{equation}
    \hat{\textbf{c}} = g_\phi(\textbf{x}, \textbf{d}, \hat{\textbf{n}}, \hat{\textbf{z}}).
\end{equation}
The 3D normal $\hat{\textbf{n}}$ is calculated as the analytical gradient of the SDF function, \ie, $\hat{\textbf{n}} = \nabla f_{\theta}(\cdot)$. The feature vector $\hat{\textbf{z}}$ is the output of a second linear head of the implicit network, as in Yariv \etal~\cite{yariv2020multiview} and Yu \etal~\cite{yu2022monosdf}.

\subsection{Supervision and learning curriculum}

We employ neural implicit shape and radiance field representation to effectively learn the 3D shape prior and to capture pixel-level details, benefiting from both explicit 3D and volume rendering supervision.

\paragraph{3D supervision}

We apply direct 3D supervision by using the following loss between the predicted and real \ac{sdf} values:
\begin{equation}
    \mathcal{L}_{3D} = \sum_{\textbf{x} \in \{\mathcal{X} \bigcup \textbf{r}\}} ||s(\textbf{x}) - \hat{s}(\textbf{x})||_1 .
\end{equation}
This loss is computed for points along the rays $\textbf{r}$ and from the point set $\mathcal{X}$, which contains uniformly sampled points and near-surface points.

\paragraph{Photometric reconstruction loss}

For all rays $\textbf{r}$ in the minibatch, we render each pixel with the predicted \ac{sdf} values $\hat{s}$ and color value $\hat{\textbf{c}}$ for all sampled points on the ray; the volume rendering formulations are detailed in \cref{sec:background}. The photometric reconstruction loss is defined as:
\begin{equation}
    \mathcal{L}_{rgb} = \sum_{\textbf{r}} ||C(\textbf{r}) - \hat{C}(\textbf{r})||_1,
\end{equation}
where $C(\textbf{r})$ denotes the color value in the input image.

\paragraph{Exploiting monocular geometric cues}

To further alleviate ambiguities in recovering 3D shapes from single-view inputs, we follow Yu \etal~\cite{yu2022monosdf} and exploit monocular depth and normal cues to facilitate the training process. The depth and normal consistency losses are defined as follows: 
\begin{equation}
    \begin{aligned}
        \mathcal{L}_{d} &= \sum_{\textbf{r}} ||D(\textbf{r}) - \hat{D}(\textbf{r})||^2\\
        \mathcal{L}_{n} &= \sum_{\textbf{r}} ||N(\textbf{r}) - \hat{N}(\textbf{r})||_1 + ||1-N(\textbf{r})^\top\hat{N}(\textbf{r})||_1
    \end{aligned}
\end{equation}
Compared to the photometric reconstruction loss, the depth and normal consistency losses can directly supervise the \ac{sdf} prediction in the implicit network $f_\theta$ through back-propagation without the rendering network $g_\phi$; please refer to \cref{fig:framework} for detailed illustration.

\paragraph{Overall loss}

The overall loss used to optimize the implicit and rendering networks jointly is given by:
\begin{equation}
    \mathcal{L} = \alpha_{3D}\mathcal{L}_{3D} + \alpha_{rgb}\mathcal{L}_{rgb} + \alpha_{d}\mathcal{L}_{d} + \alpha_{n}\mathcal{L}_{n},
\end{equation}
where $\alpha$ denotes the respective loss weight. $\mathcal{L}_{rgb}$, $\mathcal{L}_{d}$, and $\mathcal{L}_{n}$ are applied to the visible pixels for the object $\mathcal{O}$, indicated by its visible mask segmentation. Note that depth, normal, and segmentation are only used during the training stage, and none are required during the inference stage, preserving the flexibility and applicability of our model.

\paragraph{Learning curriculum}

To address the limitations of naively incorporating 3D and rendering supervision in the single-view setting, which may result in realistic images but inconsistent 3D geometry due to shape-appearance ambiguity, we introduce a learning curriculum based on two empirical observations: 1) the rendering supervision should serve as an auxiliary to 3D supervision, and 2) it is more effective to first learn the overall object shape before delving into finer details. Following these, our learning curriculum comprises two stages: \stageone, which only employs 3D supervision $\mathcal{L}_{3D}$, and \stagetwo, which incorporates $\mathcal{L}_{rgb}$, $\mathcal{L}_{d}$, and $\mathcal{L}_{\textbf{n}}$ with linearly increasing loss weights:
\begin{equation}
    \alpha = \eta(\lambda - \lambda_0), \lambda > \lambda_0.
\end{equation}
$\lambda$ denotes the epoch number, $\lambda_0$ is the starting epoch of \stagetwo, and $\eta$ is the linear coefficient. $\lambda_0$ is empirically selected by observing the shape learning curves and our curriculum is crucial for performance improvement (\cref{sec:exp_obj_recon}).

\subsection{3D scene composition}\label{sec:scene_compose}

A scene can be represented by the composition of objects $\{\mathcal{O}_i, i=1, \cdots, k\}$ within it. We obtain both the 3D reconstructed geometry and the photometric rendering of the scene by composing the implicit representations of the individual objects given their 2D and 3D bounding boxes.

\paragraph{3D scene geometry}

To compose the 3D geometry of the scene, we transform each object's implicit surfaces into explicit meshes using the marching cube algorithm~\cite{lorensen1987marching}. The object meshes are then combined using the camera's extrinsic parameters and 3D object bounding boxes.

\paragraph{3D scene rendering}

To render an image of the scene, we sample points along the rays and estimate their density and color values for each individual object. Sampled points on the same ray from different objects are then grouped together to composite the colors and densities for volume rendering. This distance-aware integration ensures that only visible objects appear in the final images, as the accumulated transmittance along the ray reflects visibility.

The object composition operation offers flexibility for both reconstruction and rendering, making it applicable for holistic scene understanding and novel view synthesis with 3D scene editing, such as object rotation, translation, and compositions from different scenes.

\begin{figure*}[t!]
	\includegraphics[width=\linewidth]{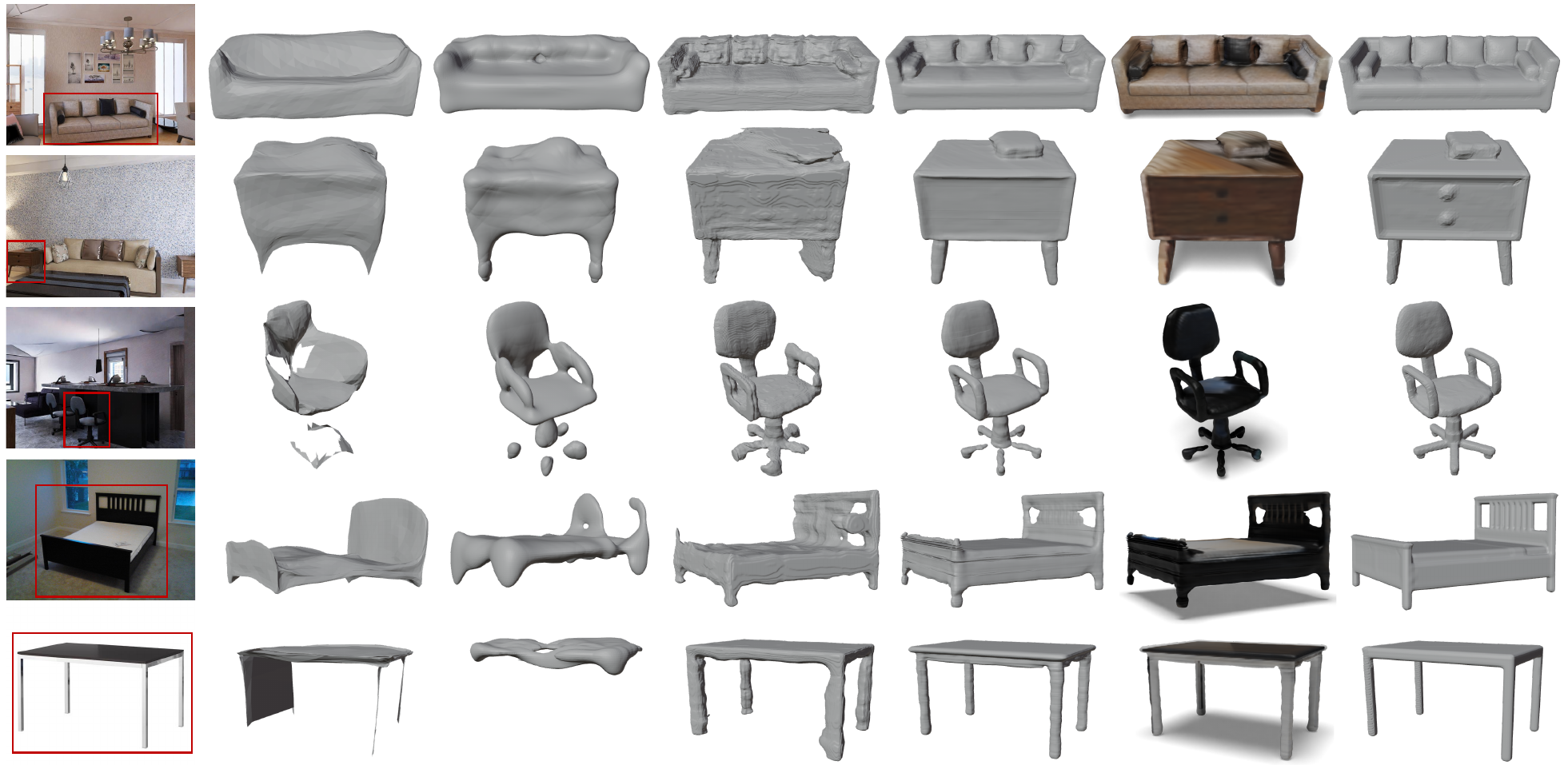}
    {\footnotesize \hspace*{0.65cm} Input \hspace{1.8cm}  MGN  \hspace{1.8cm} LIEN \hspace{1.6cm}  InstPIFu \hspace{1.3cm}  Ours$_{\text{shape}}$ \hspace{1.05cm} Ours$_{\text{shape+texture}}$ \hspace{1.15cm} GT \hfill}
	\caption{\textbf{Qualitative results of indoor object reconstruction.} Examples from \threedfront~\cite{fu20213d} (top three rows) and \pixthreed~\cite{sun2018pix3d} (bottom two rows) datasets. Our model produces textured 3D objects with smoother surfaces and finer details than previous methods.}
	\label{fig:qual_front3d}
\end{figure*}

\section{Experiment}

We evaluate single-view object reconstruction in indoor scenes using synthetic dataset \threedfront~\cite{fu20213d} and real dataset \pixthreed~\cite{sun2018pix3d}. Our model's capabilities are tested in novel view synthesis, depth estimation, and normal estimation tasks, leveraging its rendering capabilities. Additionally, we showcase potential applications of our model, including holistic scene understanding and 3D scene editing.

\subsection{Indoor object reconstruction}\label{sec:exp_obj_recon}

\paragraph{Datasets}

We evaluate our single-view object reconstruction on synthetic dataset \threedfront~\cite{fu20213d} and real dataset \pixthreed~\cite{sun2018pix3d}. We adopt the same splits as Liu \etal~\cite{liu2022towards} for both datasets. Data preparation details, including monocular cues and \ac{sdf} generation, are in \supmat.

\paragraph{Evaluation metrics}

To evaluate 3D object reconstruction, we use \ac{cd}, \fscore, and \ac{nc} following Wang \etal~\cite{wang2018pixel2mesh} and Mescheder \etal~\cite{mescheder2019occupancy}. \ac{cd} measures the sum of squared distances between the nearest neighbor correspondences of two point clouds after mesh alignment. \fscore~\cite{knapitsch2017tanks} is the harmonic mean of precision and recall of points in the prediction and ground truth within the nearest neighbor. \ac{nc} quantifies how well methods capture higher-order information by computing the mean absolute dot product of the normals between meshes after alignment.

\paragraph{Results}

In the single-view object reconstruction task, we compare our method with MGN of Total3D~\cite{nie2020total3d}, the LIEN of Im3D~\cite{zhang2021holistic}, and InstPIFu~\cite{liu2022towards}. Our model surpasses \sota methods across all three metrics in both synthetic (\cref{tab:quant_front3d}) and real (\cref{tab:quant_pix3d}) datasets. Notably, on \threedfront, our model achieves $27.7\%$ and $16.4\%$ performance gain in mean \ac{cd} and \fscore, respectively, as well as the best \ac{nc} on all object categories. This highlights our model's proficiency in predicting highly detailed object shapes and smoother surfaces (\cref{fig:qual_front3d}). Moreover, our model predicts high-fidelity textures, a capability lacking in previous models. It represents significant progress in single-view 3D object reconstruction, enabling the recovery of both fine-grained shapes and intricate textures. Further details, results, and failure cases are in \supmat.

\begin{table}[b!]
    \centering
    \small
    \setlength{\tabcolsep}{3pt}
    \caption{\textbf{Object reconstruction on the \threedfront~\cite{fu20213d} dataset.} Our model achieves the best performance on mean \acs{cd} and \fscore, as well as the best \acs{nc} on all object categories, outperforming MGN~\cite{nie2020total3d}, LIEN~\cite{zhang2021holistic}, and InstPIFu~\cite{liu2022towards}. $\dagger$: Results reproduced from the official repository.}
    \resizebox{\linewidth}{!}{%
        \begin{tabular}{ccccccccccc}
            \toprule
            \multicolumn{2}{c}{\textbf{Category}} & bed & chair & sofa & table & desk & nightstand & cabinet & bookshelf & mean \\
            \midrule
            \multirow{4}{*}{\textbf{\acs{cd} $\downarrow$}} & MGN  & 15.48  & 11.67  & 8.72  & 20.90  & 17.59  & 17.11  & 13.13  & 10.21  & 14.07  \\
            & LIEN & 16.81  & 41.40  & 9.51  & 35.65  & 26.63  & 16.78  & 7.44  & 11.70  & 28.52  \\
            & InstPIFu & 18.17  & 14.06  & 7.66  & 23.25  & 33.33  & \textbf{11.73}  & \textbf{6.04}  & 8.03  & 14.46  \\
            & Ours & \textbf{4.96}  & \textbf{10.52}  & \textbf{4.53}  & \textbf{16.12}  & \textbf{25.86}  & 17.90  & 6.79  & \textbf{3.89}  & \textbf{10.45}  \\
            \midrule
            \multirow{4}{*}{\textbf{$\fscore\uparrow$}} & MGN & 46.81  & 57.49  & 64.61  & 49.80  & 46.82  & 47.91  & 54.18  & 54.55  & 55.64  \\
            & LIEN & 44.28  & 31.61  & 61.40  & 43.22  & 37.04  & 50.76  & 69.21  & 55.33  & 45.63  \\
            & InstPIFu & 47.85  & 59.08  & 67.60  & 56.43  & \textbf{48.49}  & 57.14  & \textbf{73.32}  & 66.13  & 61.32  \\
            & Ours & \textbf{76.34}  & \textbf{69.17}  & \textbf{80.06}  & \textbf{67.29}  & 47.12  & \textbf{58.48}  & 70.45  & \textbf{85.93}  & \textbf{71.36}  \\
            \midrule
            \multirow{4}{*}{\textbf{$\acs{nc}\uparrow$}}     & MGN$^\dagger$ & 0.829  & 0.758  & 0.819  & 0.785  & 0.711  & 0.833  & 0.802  & 0.719  & 0.787  \\
            & LIEN$^\dagger$ & 0.822  & 0.793  & 0.803  & 0.755  & 0.701  & 0.814  & 0.801  & 0.747  & 0.786  \\
            & InstPIFu$^\dagger$ & 0.799  & 0.782  & 0.846  & 0.804  & 0.708  & 0.844  & 0.841  & 0.790  & 0.810  \\
            & Ours & \textbf{0.896}  & \textbf{0.833}  & \textbf{0.894}  & \textbf{0.838}  & \textbf{0.764}  & \textbf{0.897}  & \textbf{0.856}  & \textbf{0.862}  & \textbf{0.854}  \\ 
            \bottomrule
        \end{tabular}%
    }%
    \label{tab:quant_front3d}
\end{table}

\begin{table}[b!]
    \centering
    \small
    \setlength{\tabcolsep}{3pt}
    \caption{\textbf{Object Reconstruction on the \pixthreed~\cite{sun2018pix3d} dataset.} On the non-overlapped split~\cite{liu2022towards}, our model outperforms the state-of-the-art methods by significant margins. $\dagger$: Results reproduced from the official repository.}
    \resizebox{\linewidth}{!}{%
        \begin{tabular}{lcccccccccccc}
            \toprule
            \multicolumn{2}{c}{\textbf{Category}} & bed & bookcase & chair & desk & sofa & table & tool & wardrobe & misc & mean \\
            \midrule
            \multirow{4}{*}{\textbf{\acs{cd} $\downarrow$}} & MGN & 22.91 & 33.61 & 56.47 & 33.95 & 9.27 & 81.19 & 94.70 & 10.43 & 137.50 & 44.32 \\
            & LIEN & 11.18 & 29.61 & 40.01 & 65.36 & 10.54 & 146.13 & 29.63 & 4.88 & 144.06 & 51.31 \\
            & InstPIFu & 10.90 & 7.55 & 32.44 & \textbf{22.09} & 8.13 & 45.82 & 10.29 & \textbf{1.29} & 47.31 & 24.65 \\
            & Ours & \textbf{6.31} & \textbf{7.21} & \textbf{26.23} & 28.63 & \textbf{5.68} & \textbf{43.87} & \textbf{8.29} & 2.07 & \textbf{35.03} & \textbf{21.79} \\
            \midrule
            \multirow{4}{*}{\textbf{$\fscore\uparrow$}} & MGN & 34.69 & 28.42 & 35.67 & 65.36 & 51.15 & 17.05 & 57.16 & 52.04 & 10.41 & 36.20 \\
            & LIEN & 37.13 & 15.51 & 25.70 & 26.01 & 49.71 & 21.16 & 5.85 & 59.46 & 11.04 & 31.45 \\
            & InstPIFu & 54.99 & 62.26 & 35.30 & \textbf{47.30} & 56.54 & 37.51 & 64.24 & \textbf{94.62} & 27.03 & 45.62 \\
            & Ours & \textbf{68.78} & \textbf{66.69} & \textbf{55.18} & 42.49 & \textbf{71.22} & \textbf{51.93} & \textbf{65.38} & 91.84 & \textbf{46.92} & \textbf{59.71} \\
            \midrule
            \multirow{4}{*}{\textbf{$\acs{nc}\uparrow$}} & MGN$^\dagger$ & 0.737 & 0.592 & 0.525 & 0.633 & 0.756 & 0.794 & 0.531 & 0.809 & 0.563 & 0.659 \\
            & LIEN$^\dagger$ & 0.706 & 0.514 & 0.591 & 0.581 & 0.775 & 0.619 & 0.506 & 0.844 & 0.481 & 0.646 \\
            & InstPIFu$^\dagger$ & 0.782 & 0.646 & 0.547 & 0.758 & 0.753 & 0.796 & 0.639 & 0.951 & 0.580 & 0.683 \\
            & Ours & \textbf{0.825} & \textbf{0.689} & \textbf{0.693} & \textbf{0.776} & \textbf{0.866} & \textbf{0.835} & \textbf{0.645} & \textbf{0.960} & \textbf{0.599} & \textbf{0.778} \\
            \bottomrule
        \end{tabular}%
    }%
    \label{tab:quant_pix3d}
\end{table}

\begin{table}[ht!]
    \centering
    \small
    \setlength{\tabcolsep}{7pt}
    \caption{\textbf{Ablation studies on object reconstruction.} We demonstrate the benefits of introducing 2D supervision and employing a properly designed curriculum. The notation ${\times5/10}$ indicates increased loss weights.}
    \resizebox{\linewidth}{!}{%
        \begin{tabular}{lllllccc}
            \toprule
            $\ac{sdf}$        &        $C$       &       $D$       &       $N$       &       $Curr.$         & $\acs{cd}\downarrow$ & $\fscore\uparrow$ & $\acs{nc}\uparrow$ \\
            \midrule 
            \checkmark & $\times$       & $\times$      & $\times$      & $\times$            & 16.43          & 64.15          & 0.806          \\
            \checkmark & \checkmark     & $\times$      & $\times$      & $\times$            & 15.90          & 65.55          & 0.813          \\
            \checkmark & \checkmark     & \checkmark    & $\times$      & $\times$            & 14.02          & 67.31          & 0.828          \\
            \checkmark & \checkmark     & \checkmark    & \checkmark    & $\times$            & 12.92          & 67.71          & 0.841          \\
            \checkmark & \checkmark$_{\times5}$    & \checkmark$_{\times5}$    & \checkmark$_{\times5}$    & $\times$            & 16.19         & 64.92         & 0.813         \\
            \checkmark & \checkmark$_{\times10}$    & \checkmark$_{\times10}$    & \checkmark$_{\times10}$    & $\times$            & 19.22         & 58.26         & 0.771         \\
            \checkmark & \checkmark     & \checkmark    & \checkmark    & $\lambda_0=0$         & 12.88          & 68.19          & 0.840          \\
            \checkmark & \checkmark     & \checkmark    & \checkmark    & $\lambda_0=70$        & 12.42          & 68.87          & 0.845          \\
            \checkmark & \checkmark     & \checkmark    & \checkmark    & $\lambda_0=150$       & \textbf{10.45} & \textbf{71.36} & \textbf{0.854} \\
            \bottomrule  
        \end{tabular}%
    }%
    \label{tab:ablation}
\end{table}

\paragraph{Ablations}

\begin{figure}[t!]
	\includegraphics[width=\linewidth]{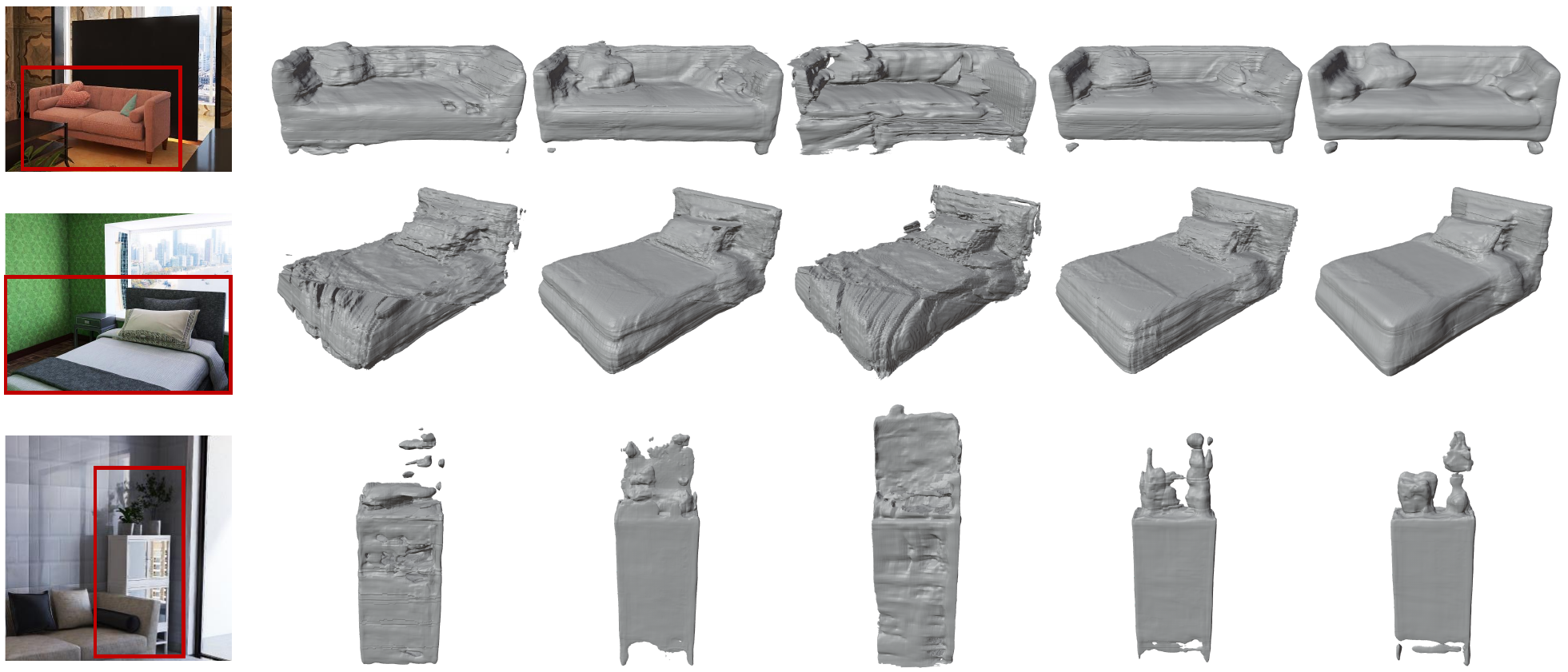}
    {\footnotesize \hspace*{0.2cm} Input \hspace{0.95cm}   (a) \hspace{0.9cm} (b) \hspace{0.95cm}  (c) \hspace{0.92cm}   (d) \hspace{0.9cm} (e) \hfill }
	\caption{\textbf{Visual comparisons for ablation study.} (a) $\ac{sdf}$ only (b) $\ac{sdf}+C+D+N$ (c) $\ac{sdf}+C+D+N$ with ${\times10}$ loss weights \checkmark$_{\times10}$ (d) $\lambda_0=0$ (e) $\lambda_0=150$. Incorporating 2D supervision with our designed curriculum yields the best reconstruction quality.}
	\label{fig:ablation}
\end{figure}

To analyze the effects of various supervisions and learning curricula, we conduct ablative studies on $\ac{sdf}$, color ($C$), depth ($D$), and normal ($N$) supervisions, along with different loss weights and curriculum strategies ($Curr.$). Key findings from \cref{tab:ablation,fig:ablation} are as follows:
\begin{enumerate}[leftmargin=*]
    \item Incorporating color, depth, and normal supervision in our framework significantly enhances 3D object reconstruction, especially in capturing finer details.
    \item 2D supervision should act as an auxiliary to 3D supervision, as simply increasing 2D loss weights (\eg, $\checkmark_{\times5}$ or $\checkmark_{\times10}$) negatively impacts 3D reconstruction. \cref{fig:ablation}{\color{red}(c)} shows clear artifacts along the ray directions, indicating the importance of a suitable learning curriculum. 
    \item Our proposed curriculum, gradually increasing 2D loss weights after the 3D shape prior learning phase (\stagetwo starting epoch $\lambda_0=150$), outperforms early injection of 2D supervision ($\lambda_0=0$ or $\lambda_0=70$). 
\end{enumerate}

\begin{figure}[t!]
	\centering
	\includegraphics[width=\linewidth]{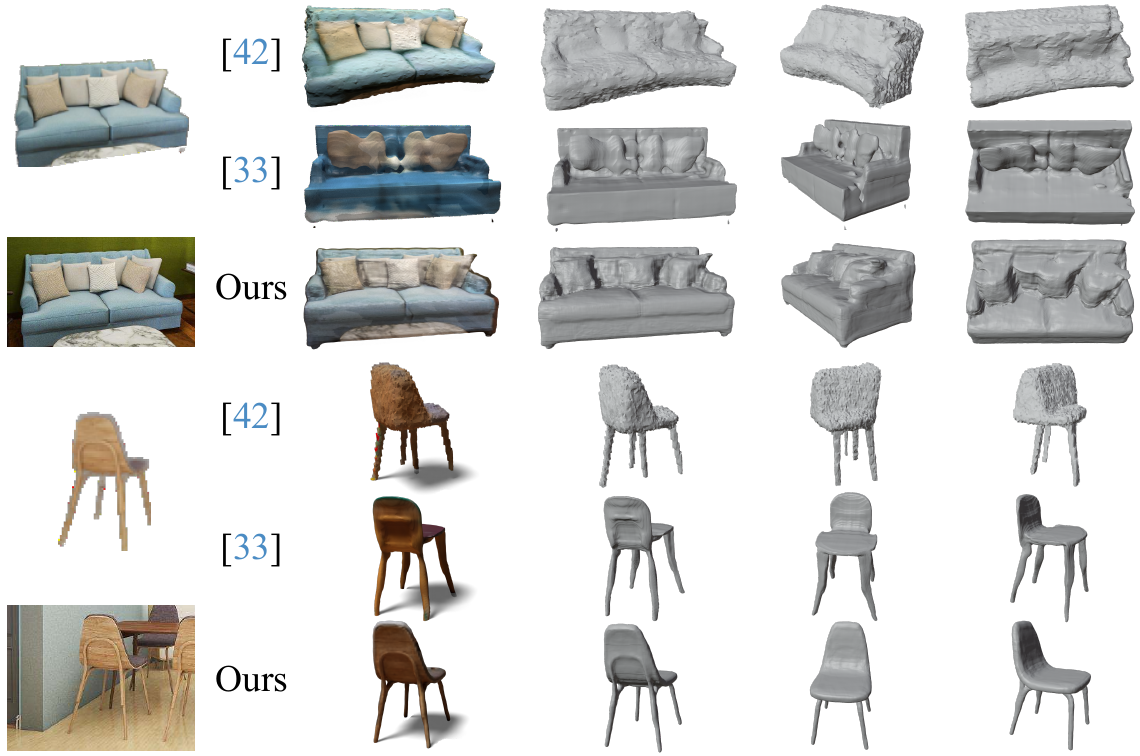}
	\caption{\textbf{Comparison with prior-guided models.} Inputs for Zero-1-to-3~\cite{liu2023zero1to3} and \shapee~\cite{jun2023shape} only contains foreground objects. Each example is presented with textured mesh and mesh from three views. Our method outperforms prior-guided models in capturing details and 3D shape consistency.}
	\label{fig:qual_prior}
\end{figure}

\paragraph{Comparison with prior-guided models}

We compare our model with generative models demonstrating potential zero-shot generalizability by leveraging 2D or 3D geometric priors learned from \textit{large-scale} datasets. Specifically, we choose two representative works: (1) Zero-1-to-3~\cite{liu2023zero1to3}, which uses Objaverse~\cite{deitke2023objaverse} to learn a 2D diffusion prior for novel view synthesis under specified camera transformation and reconstructs objects under a multi-view setting; (2) \shapee~\cite{jun2023shape}, which directly generates textured meshes given images and category prompts, trained on millions of paired 3D and text data. 
For a fair evaluation, we compare them with our model on a subset of the test split in \threedfront with ground truth object scale. Results in \cref{fig:qual_prior,tab:quant-prior} show that while Zero-1-to-3 produces reasonable images on particular views, it faces difficulties achieving overall 3D shape consistency. \shapee captures the rough object shape but lacks detailed modeling of geometry and texture. On the contrary, our model excels at recovering the general 3D shapes while maintaining fine geometrical and textural details. This stresses the significant potential of effectively integrating 2D and 3D priors for future single-view reconstruction models to achieve enhanced results and generalizability. More details can be found in \supmat.

\begin{table}[t!]
    \centering
    \small
    \setlength{\tabcolsep}{3pt}
    \caption{\textbf{Quantitative comparison with prior-guided models.} Despite the zero-shot generalization ability, methods leveraging 2D or 3D priors fall short in recovering object geometry, especially surface details, compared to our proposed method.}
    \begin{tabular}{lccc}
        \toprule
         & $\acs{cd}\downarrow$ & $\fscore\uparrow$ & $\acs{nc}\uparrow$\\
        \midrule
        Zero-1-to-3~\cite{liu2023zero1to3} & 39.27 & 30.07 & 0.624\\
        \shapee~\cite{jun2023shape} & 29.16 & 39.86 & 0.686\\
        Ours & \textbf{10.86} & \textbf{69.95} & \textbf{0.846}\\
        \bottomrule
    \end{tabular}%
    \label{tab:quant-prior}
\end{table}

\begin{figure}[t!]
	\centering
	\includegraphics[width=\linewidth]{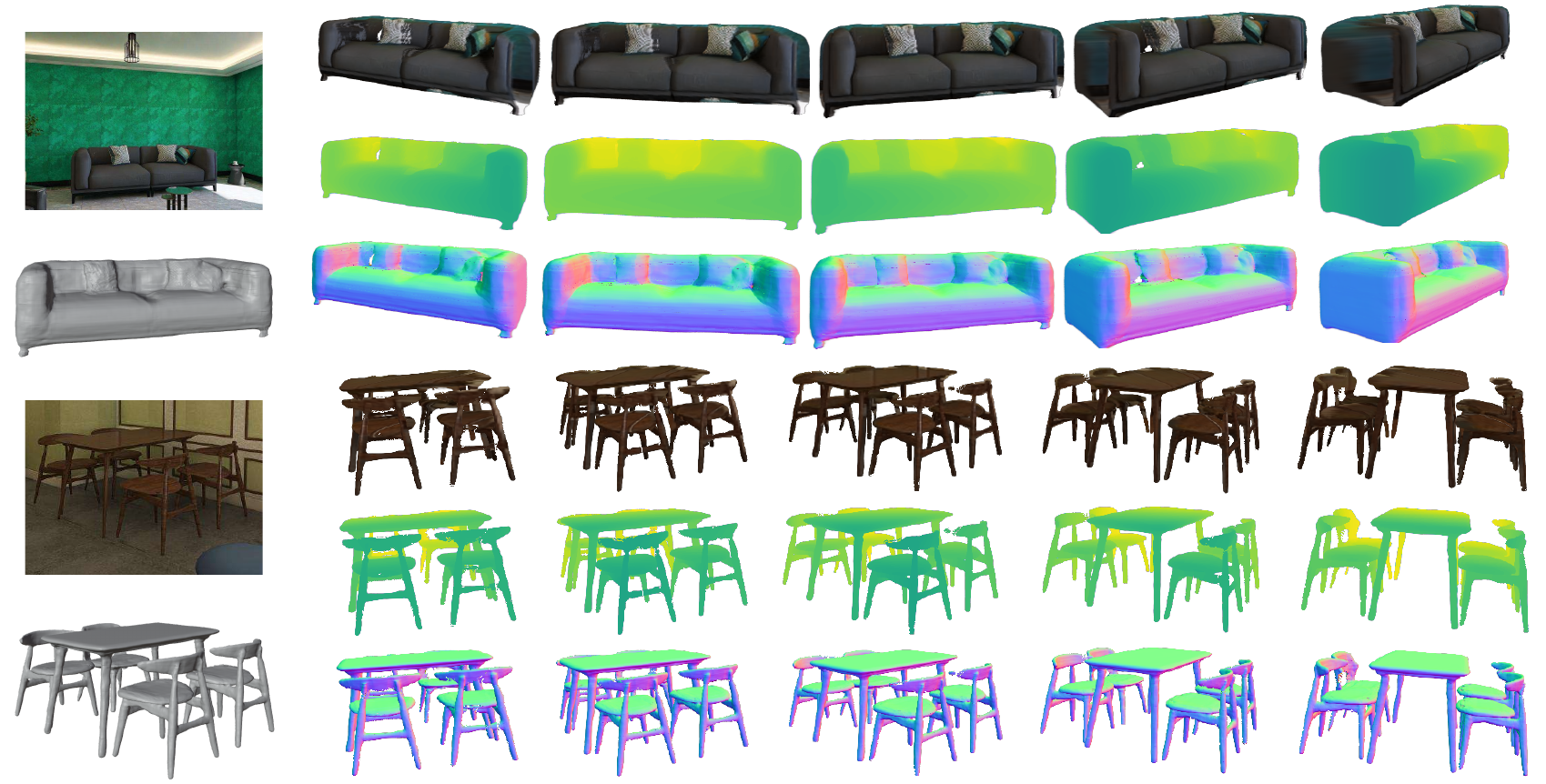}
    {\footnotesize \hspace*{0.1cm} Input\&Rec. \hspace{0.50cm} -40\textdegree \hspace{0.65cm} -20\textdegree \hspace{0.88cm}  0\textdegree \hspace{0.95cm} 20\textdegree \hspace{0.9cm} 40\textdegree \hfill}
	\caption{\textbf{Novel view rendering from single-view inputs.} Our model can render color, depth, and normal images for both objects (top) and scenes (bottom) from novel views.}
	\label{fig:render}
\end{figure}

\subsection{Rendering capability}

Harnessing the advantages of our method, we seamlessly introduce rendering capabilities to a single-view reconstruction model. From the single-view input image, we can render the color, depth, and normal images through volume rendering, even from novel views. The qualitative examples presented in \cref{fig:render} illustrate that our method excels in producing plausible and consistent rendering results, even when the viewing angles change significantly (\ie, $\pm40$\textdegree).

\begin{figure}[t!]
	\centering
	\includegraphics[width=\linewidth]{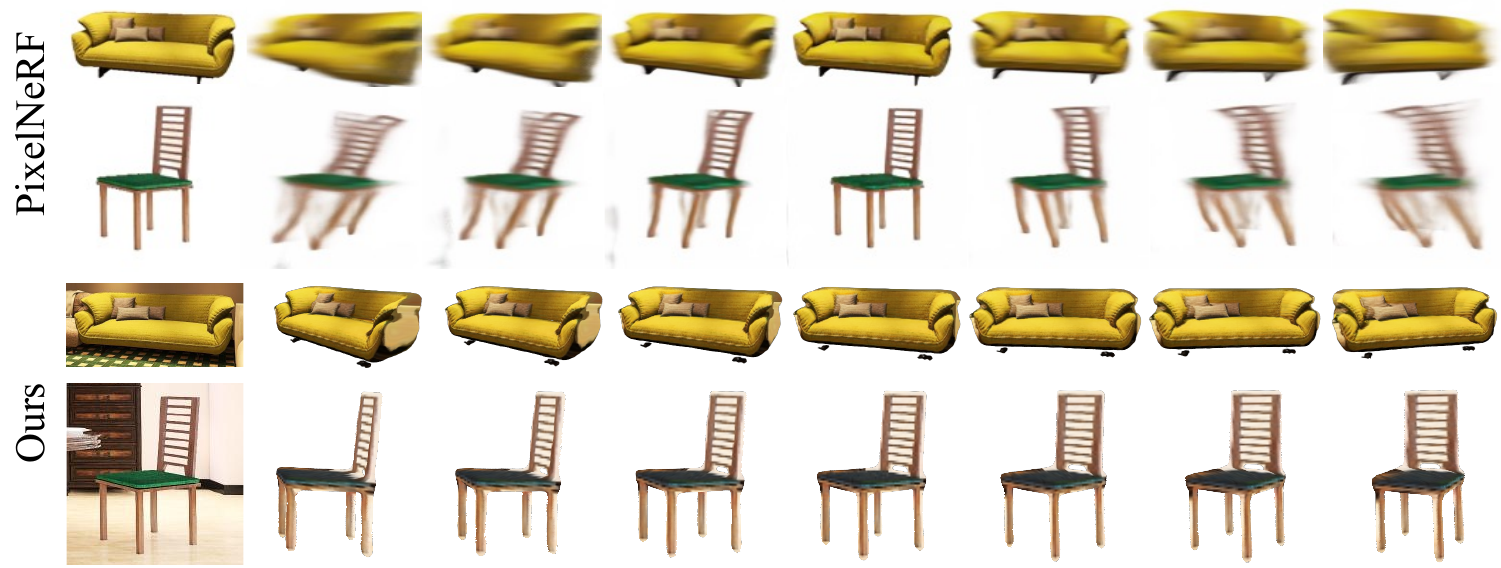}
    {\footnotesize \hspace*{0.53cm} Input \hspace{0.3cm} -30\textdegree \hspace{0.43cm} -20\textdegree \hspace{0.45cm}  -10\textdegree \hspace{0.56cm} 0\textdegree \hspace{0.60cm} 10\textdegree \hspace{0.65cm} 20\textdegree \hspace{0.5cm} 30\textdegree \hfill}
	\caption{\textbf{Visual comparison for novel view rendering.} PixelNeRF~\cite{yu2021pixelnerf} (top two rows) struggles to render images outside the vicinity of the input view, whereas our method (bottom two rows) can produce realistic renderings even for views far from the original input. PixelNeRF inputs only contain foreground objects.}
	\label{fig:pixelnerf}
\end{figure}

\paragraph{Novel view synthesis}

PixelNeRF~\cite{yu2021pixelnerf} employs a NeRF representation for novel view synthesis from input images.  We compare the class-agnostic model of PixelNeRF, which is pre-trained on ShapeNet~\cite{shapenet2015} and fine-tuned on \threedfront.  Qualitative results in \cref{fig:pixelnerf} reveal a notable difference between the two approaches: PixelNeRF struggles to render images outside the vicinity of the original viewpoints, whereas our method is capable of generating meaningful renderings from novel viewpoints. This shows the importance of effectively imposing explicit 3D shapes in the scene reconstruction model, particularly when dealing with partial observation in real scenes.

\paragraph{Depth and normal estimation}

Moreover, our model can serve as a proficient single-view depth and normal estimator. To validate this, we compare with zero-shot \sota methods~\cite{zamir2020robust,eftekhar2021omnidata} on \threedfront, following Ranftl \etal~\cite{ranftl2021vision}. Results in \cref{tab:depth_normal} demonstrate that our model performs comparably on the input views. It also shows our model can directly estimate reasonable depth and normal maps on novel views. This is challenging since our model solely relies on single-view inputs, which is in stark contrast from previous work~\cite{yu2022monosdf,guo2022neural,wang2022neuris} that require multi-view inputs; see \supmat for additional results. 

\begin{table}[t!]
    \centering
    \small
    \caption{\textbf{Single-view depth and normal estimation.} We evaluate depth using L1 $\downarrow$ and normal using L1$\downarrow$ / Angular$^{\circ}$$\downarrow$ error as metrics. For novel views, we use $\pm15$\textdegree \xspace views to evaluate the accuracy.}
    \resizebox{\linewidth}{!}{%
        \begin{tabular}{lcccccc}
            \toprule
            & \multicolumn{2}{c}{Original View} & \multicolumn{2}{c}{Novel View} \\
            \cmidrule(lr){2-3} \cmidrule(lr){4-5}
            & Depth & Normal & Depth & Normal\\
            \cmidrule(lr){2-5}
            XTC~\cite{zamir2020robust} & 1.188 & 12.712 / 14.309 & - & - \\
            Omnidata~\cite{eftekhar2021omnidata} & \textbf{0.734} & \textbf{10.015} / \textbf{11.257} & -         & - \\
            Ours & 0.992 & 10.962 / 12.392 & \textbf{1.179} & \textbf{12.094} / \textbf{13.710} \\
            \bottomrule
        \end{tabular}%
    }%
    \label{tab:depth_normal}
\end{table}

\subsection{Applications}

\begin{figure}[t!]
	\includegraphics[width=\linewidth]{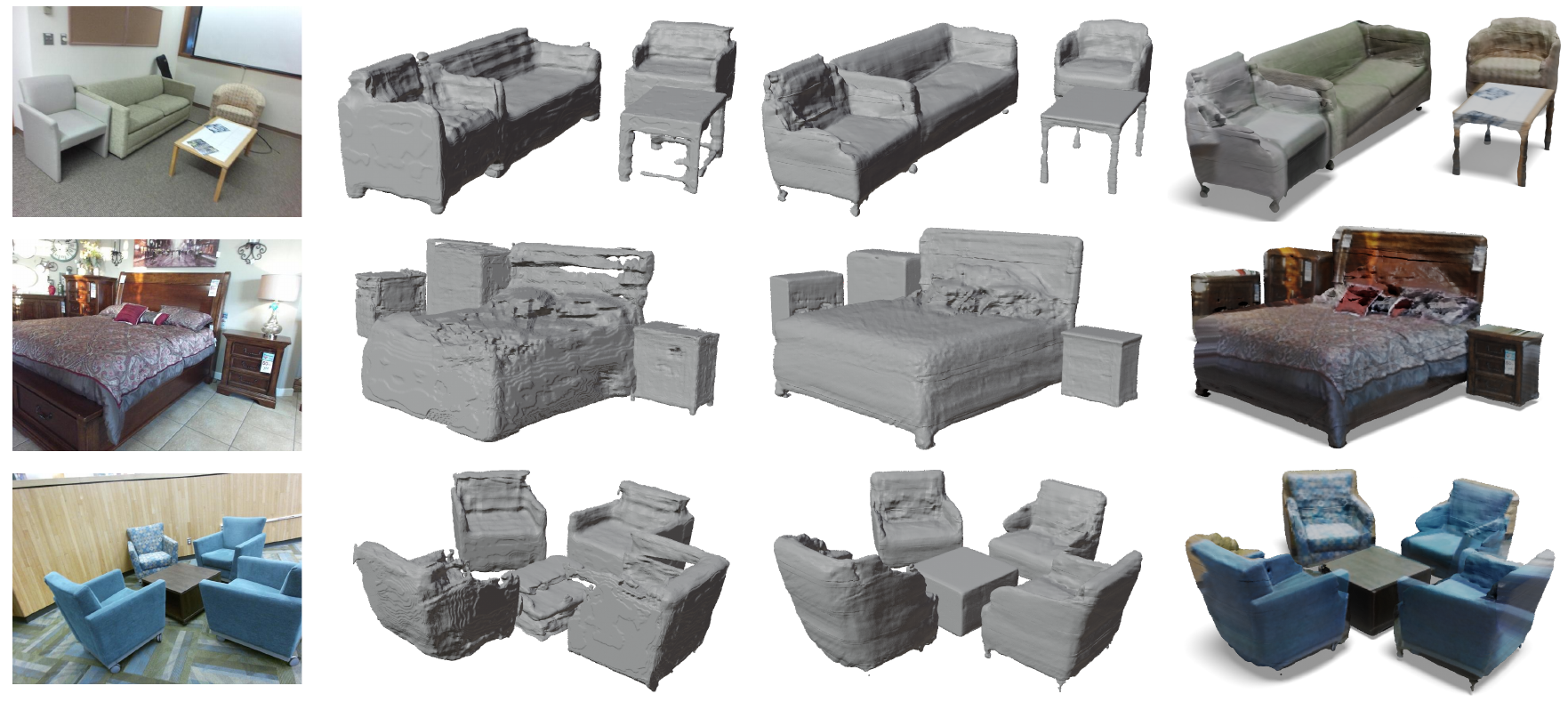}
    {\footnotesize \hspace*{0.47cm} Input \hspace{1.10cm} InstPIFU \hspace{1.0cm} Ours$_{\text{shape}}$ \hspace{0.7cm}  Ours$_{\text{shape+texture}}$  \hfill}
	\caption{\textbf{Holistic scene understanding and generalization.} The reconstruction results on \sunrgbd~\cite{song2015sun} with existing 3D object detectors demonstrate our model's performance in recovering realistic scenes with generalization ability.}
	\label{fig:holistic_scene}
\end{figure}

\begin{figure}[t!]
	\centering
	\includegraphics[width=\linewidth]{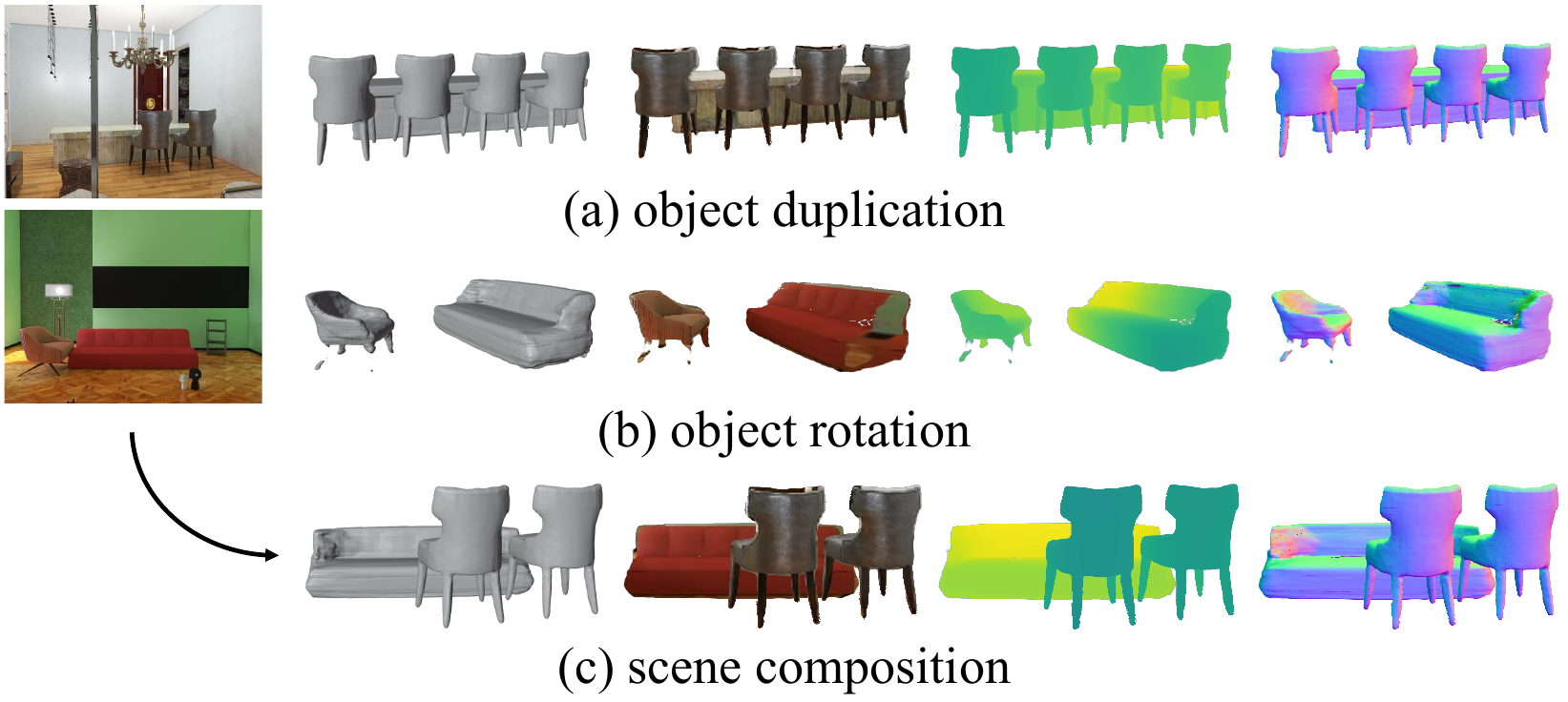}
	\caption{\textbf{3D scene editing based on single-view inputs.} Reconstructions and renderings are shown for (a) duplicating the chairs, (b) rotating the sofa and (c) scene composition of (a) and (b).}
	\label{fig:scene_edit}
\end{figure}

\paragraph{Generalizable holistic scene understanding}

Our method is capable of recovering 3D scene geometry and rendering corresponding color, depth, and normal images by composing object-level implicit representations (see \cref{sec:scene_compose} for more details). \cref{fig:holistic_scene} showcases qualitative scene reconstruction results on \sunrgbd~\cite{song2015sun} by employing existing 3D object detectors~\cite{zhang2021holistic,brazil2023omni3d}. The results demonstrate that \textit{our method can reconstruct detailed object shapes and intricate textures in real images with cross-domain generalization ability}.

\paragraph{Scene editing}

Finally, we demonstrate our model's potential in representing scenes and enabling 3D scene editing applications. Our method allows for object-level editing, such as object translation, rotation, duplication, and composition of objects from different scenes into a shared 3D space. Qualitative results are shown in \cref{fig:scene_edit}. Notably, \textit{our approach can generate both 3D geometry and rendered images for edited scenes}, which differentiates itself from previous work that could only render images of manipulated objects~\cite{niemeyer2021giraffe,wu2022object,yang2021learning}, perform color or texture editing~\cite{lazova2023control}, or require multi-view posed images as input~\cite{liu2020neural,yuan2022nerf}.

\section{Conclusion}

We present a novel framework for single-view scene reconstruction utilizing neural implicit shape and radiance field representations. Our model exhibits a significant advantage in textured 3D object reconstruction compared to \sota methods, and integrating color, depth, and normal supervision with our designed curriculum is pivotal to achieving improved performance. Furthermore, our model demonstrates impressive rendering capabilities and performs well in single-view depth and normal estimation, showing promise for generalization in holistic scene understanding and facilitating applications like 3D scene editing. Potential limitations include the model's ability to reconstruct objects from novel categories and texture recovery for unseen object parts. Effectively incorporating 2D or 3D priors from large-scale datasets offers a promising avenue for future direction.

\paragraph{Acknowledgment}

The authors thank colleagues from BIGAI for fruitful discussions and anonymous reviewers for constructive feedback. This work is supported in part by the National Key R\&D Program of China (2021ZD0150200).

{
    \small
    \bibliographystyle{ieeenat_fullname}
    \bibliography{reference_header,reference}

\begin{thebibliography}{86}
\providecommand{\natexlab}[1]{#1}
\providecommand{\url}[1]{\texttt{#1}}
\expandafter\ifx\csname urlstyle\endcsname\relax
  \providecommand{\doi}[1]{doi: #1}\else
  \providecommand{\doi}{doi: \begingroup \urlstyle{rm}\Url}\fi

\bibitem[Achlioptas et~al.(2018)Achlioptas, Diamanti, Mitliagkas, and Guibas]{achlioptas2018learning}
Panos Achlioptas, Olga Diamanti, Ioannis Mitliagkas, and Leonidas Guibas.
\newblock Learning representations and generative models for 3d point clouds.
\newblock In \emph{International Conference on Machine Learning (ICML)}, 2018.

\bibitem[Achlioptas et~al.(2020)Achlioptas, Abdelreheem, Xia, Elhoseiny, and Guibas]{achli2020referit3d}
Panos Achlioptas, Ahmed Abdelreheem, Fei Xia, Mohamed Elhoseiny, and Leonidas Guibas.
\newblock Referit3d: Neural listeners for fine-grained 3d object identification in real-world scenes.
\newblock In \emph{European Conference on Computer Vision (ECCV)}, 2020.

\bibitem[Azuma et~al.(2022)Azuma, Miyanishi, Kurita, and Kawanabe]{azuma2022scanqa}
Daichi Azuma, Taiki Miyanishi, Shuhei Kurita, and Motoaki Kawanabe.
\newblock Scanqa: 3d question answering for spatial scene understanding.
\newblock In \emph{Conference on Computer Vision and Pattern Recognition (CVPR)}, 2022.

\bibitem[Besl and McKay(1992)]{besl1992method}
Paul~J Besl and Neil~D McKay.
\newblock Method for registration of 3-d shapes.
\newblock In \emph{Sensor fusion IV: Control paradigms and data structures}, 1992.

\bibitem[Bokhovkin et~al.(2023)Bokhovkin, Tulsiani, and Dai]{bokhovkin2023mesh2tex}
Alexey Bokhovkin, Shubham Tulsiani, and Angela Dai.
\newblock Mesh2tex: Generating mesh textures from image queries.
\newblock In \emph{International Conference on Computer Vision (ICCV)}, 2023.

\bibitem[Brazil et~al.(2023)Brazil, Kumar, Straub, Ravi, Johnson, and Gkioxari]{brazil2023omni3d}
Garrick Brazil, Abhinav Kumar, Julian Straub, Nikhila Ravi, Justin Johnson, and Georgia Gkioxari.
\newblock Omni3d: A large benchmark and model for 3d object detection in the wild.
\newblock In \emph{Conference on Computer Vision and Pattern Recognition (CVPR)}, 2023.

\bibitem[Chang et~al.(2015)Chang, Funkhouser, Guibas, Hanrahan, Huang, Li, Savarese, Savva, Song, Su, Xiao, Yi, and Yu]{shapenet2015}
Angel~X. Chang, Thomas Funkhouser, Leonidas Guibas, Pat Hanrahan, Qixing Huang, Zimo Li, Silvio Savarese, Manolis Savva, Shuran Song, Hao Su, Jianxiong Xiao, Li Yi, and Fisher Yu.
\newblock Shapenet: An information-rich 3d model repository.
\newblock \emph{arXiv preprint arXiv:1512.03012}, 2015.

\bibitem[Chen et~al.(2020)Chen, Chang, and Nie{\ss}ner]{chen2020scanrefer}
Dave~Zhenyu Chen, Angel~X Chang, and Matthias Nie{\ss}ner.
\newblock Scanrefer: 3d object localization in rgb-d scans using natural language.
\newblock In \emph{European Conference on Computer Vision (ECCV)}, 2020.

\bibitem[Chen et~al.(2019)Chen, Huang, Yuan, Qi, Zhu, and Zhu]{chen2019holistic++}
Yixin Chen, Siyuan Huang, Tao Yuan, Siyuan Qi, Yixin Zhu, and Song-Chun Zhu.
\newblock Holistic++ scene understanding: Single-view 3d holistic scene parsing and human pose estimation with human-object interaction and physical commonsense.
\newblock In \emph{International Conference on Computer Vision (ICCV)}, 2019.

\bibitem[Chen and Zhang(2019)]{chen2019learning}
Zhiqin Chen and Hao Zhang.
\newblock Learning implicit fields for generative shape modeling.
\newblock In \emph{Conference on Computer Vision and Pattern Recognition (CVPR)}, 2019.

\bibitem[Chen et~al.(2021)Chen, Gholami, Nie{\ss}ner, and Chang]{chen2021scan2cap}
Zhenyu Chen, Ali Gholami, Matthias Nie{\ss}ner, and Angel~X Chang.
\newblock Scan2cap: Context-aware dense captioning in rgb-d scans.
\newblock In \emph{Conference on Computer Vision and Pattern Recognition (CVPR)}, 2021.

\bibitem[Colomina and Molina(2014)]{colomina2014unmanned}
Ismael Colomina and Pere Molina.
\newblock Unmanned aerial systems for photogrammetry and remote sensing: A review.
\newblock \emph{ISPRS Journal of photogrammetry and remote sensing}, 2014.

\bibitem[Dai et~al.(2017)Dai, Chang, Savva, Halber, Funkhouser, and Nie{\ss}ner]{dai2017scannet}
Angela Dai, Angel~X Chang, Manolis Savva, Maciej Halber, Thomas Funkhouser, and Matthias Nie{\ss}ner.
\newblock Scannet: Richly-annotated 3d reconstructions of indoor scenes.
\newblock In \emph{Conference on Computer Vision and Pattern Recognition (CVPR)}, 2017.

\bibitem[Dasgupta et~al.(2016)Dasgupta, Fang, Chen, and Savarese]{dasgupta2016delay}
Saumitro Dasgupta, Kuan Fang, Kevin Chen, and Silvio Savarese.
\newblock Delay: Robust spatial layout estimation for cluttered indoor scenes.
\newblock In \emph{Conference on Computer Vision and Pattern Recognition (CVPR)}, 2016.

\bibitem[Deitke et~al.(2023)Deitke, Schwenk, Salvador, Weihs, Michel, VanderBilt, Schmidt, Ehsani, Kembhavi, and Farhadi]{deitke2023objaverse}
Matt Deitke, Dustin Schwenk, Jordi Salvador, Luca Weihs, Oscar Michel, Eli VanderBilt, Ludwig Schmidt, Kiana Ehsani, Aniruddha Kembhavi, and Ali Farhadi.
\newblock Objaverse: A universe of annotated 3d objects.
\newblock In \emph{Conference on Computer Vision and Pattern Recognition (CVPR)}, 2023.

\bibitem[Du et~al.(2018)Du, Liu, Basevi, Leonardis, Freeman, Tenenbaum, and Wu]{du2018learning}
Yilun Du, Zhijian Liu, Hector Basevi, Ales Leonardis, Bill Freeman, Josh Tenenbaum, and Jiajun Wu.
\newblock Learning to exploit stability for 3d scene parsing.
\newblock In \emph{Advances in Neural Information Processing Systems (NeurIPS)}, 2018.

\bibitem[Eftekhar et~al.(2021)Eftekhar, Sax, Malik, and Zamir]{eftekhar2021omnidata}
Ainaz Eftekhar, Alexander Sax, Jitendra Malik, and Amir Zamir.
\newblock Omnidata: A scalable pipeline for making multi-task mid-level vision datasets from 3d scans.
\newblock In \emph{Conference on Computer Vision and Pattern Recognition (CVPR)}, 2021.

\bibitem[Eslami et~al.(2018)Eslami, Jimenez~Rezende, Besse, Viola, Morcos, Garnelo, Ruderman, Rusu, Danihelka, Gregor, et~al.]{eslami2018neural}
SM~Ali Eslami, Danilo Jimenez~Rezende, Frederic Besse, Fabio Viola, Ari~S Morcos, Marta Garnelo, Avraham Ruderman, Andrei~A Rusu, Ivo Danihelka, Karol Gregor, et~al.
\newblock Neural scene representation and rendering.
\newblock \emph{Science}, 360\penalty0 (6394):\penalty0 1204--1210, 2018.

\bibitem[Fu et~al.(2021)Fu, Cai, Gao, Zhang, Wang, Li, Zeng, Sun, Jia, Zhao, et~al.]{fu20213d}
Huan Fu, Bowen Cai, Lin Gao, Ling-Xiao Zhang, Jiaming Wang, Cao Li, Qixun Zeng, Chengyue Sun, Rongfei Jia, Binqiang Zhao, et~al.
\newblock 3d-front: 3d furnished rooms with layouts and semantics.
\newblock In \emph{International Conference on Computer Vision (ICCV)}, 2021.

\bibitem[Gadre et~al.(2023)Gadre, Wortsman, Ilharco, Schmidt, and Song]{gadre2022cow}
Samir~Yitzhak Gadre, Mitchell Wortsman, Gabriel Ilharco, Ludwig Schmidt, and Shuran Song.
\newblock Cows on pasture: Baselines and benchmarks for language-driven zero-shot object navigation.
\newblock In \emph{Conference on Computer Vision and Pattern Recognition (CVPR)}, 2023.

\bibitem[Gkioxari et~al.(2019)Gkioxari, Malik, and Johnson]{gkioxari2019mesh}
Georgia Gkioxari, Jitendra Malik, and Justin Johnson.
\newblock Mesh r-cnn.
\newblock In \emph{International Conference on Computer Vision (ICCV)}, 2019.

\bibitem[Guillard et~al.(2022)Guillard, Stella, and Fua]{guillard2022meshudf}
Benoit Guillard, Federico Stella, and Pascal Fua.
\newblock Meshudf: Fast and differentiable meshing of unsigned distance field networks.
\newblock In \emph{European Conference on Computer Vision (ECCV)}, 2022.

\bibitem[Guo et~al.(2022)Guo, Peng, Lin, Wang, Zhang, Bao, and Zhou]{guo2022neural}
Haoyu Guo, Sida Peng, Haotong Lin, Qianqian Wang, Guofeng Zhang, Hujun Bao, and Xiaowei Zhou.
\newblock Neural 3d scene reconstruction with the manhattan-world assumption.
\newblock In \emph{Conference on Computer Vision and Pattern Recognition (CVPR)}, 2022.

\bibitem[Han and Zhu(2005)]{han2005bottom}
Feng Han and Song-Chun Zhu.
\newblock Bottom-up/top-down image parsing by attribute graph grammar.
\newblock In \emph{International Conference on Computer Vision (ICCV)}, 2005.

\bibitem[He et~al.(2017)He, Gkioxari, Doll{\'a}r, and Girshick]{he2017mask}
Kaiming He, Georgia Gkioxari, Piotr Doll{\'a}r, and Ross Girshick.
\newblock Mask r-cnn.
\newblock In \emph{International Conference on Computer Vision (ICCV)}, 2017.

\bibitem[Hedau et~al.(2009)Hedau, Hoiem, and Forsyth]{hedau2009recovering}
Varsha Hedau, Derek Hoiem, and David Forsyth.
\newblock Recovering the spatial layout of cluttered rooms.
\newblock In \emph{International Conference on Computer Vision (ICCV)}, 2009.

\bibitem[Hoiem et~al.(2005)Hoiem, Efros, and Hebert]{hoiem2005automatic}
Derek Hoiem, Alexei~A Efros, and Martial Hebert.
\newblock Automatic photo pop-up.
\newblock In \emph{ACM SIGGRAPH / Eurographics Symposium on Computer Animation (SCA)}, 2005.

\bibitem[Huang et~al.(2023)Huang, Mees, Zeng, and Burgard]{huang23vlmaps}
Chenguang Huang, Oier Mees, Andy Zeng, and Wolfram Burgard.
\newblock Visual language maps for robot navigation.
\newblock In \emph{International Conference on Robotics and Automation (ICRA)}, 2023.

\bibitem[Huang et~al.(2018{\natexlab{a}})Huang, Qi, Xiao, Zhu, Wu, and Zhu]{huang2018cooperative}
Siyuan Huang, Siyuan Qi, Yinxue Xiao, Yixin Zhu, Ying~Nian Wu, and Song-Chun Zhu.
\newblock Cooperative holistic scene understanding: Unifying 3d object, layout, and camera pose estimation.
\newblock In \emph{Advances in Neural Information Processing Systems (NeurIPS)}, 2018{\natexlab{a}}.

\bibitem[Huang et~al.(2018{\natexlab{b}})Huang, Qi, Zhu, Xiao, Xu, and Zhu]{huang2018holistic}
Siyuan Huang, Siyuan Qi, Yixin Zhu, Yinxue Xiao, Yuanlu Xu, and Song-Chun Zhu.
\newblock Holistic 3d scene parsing and reconstruction from a single rgb image.
\newblock In \emph{European Conference on Computer Vision (ECCV)}, 2018{\natexlab{b}}.

\bibitem[Huang et~al.(2019)Huang, Chen, Yuan, Qi, Zhu, and Zhu]{huang2019perspectivenet}
Siyuan Huang, Yixin Chen, Tao Yuan, Siyuan Qi, Yixin Zhu, and Song-Chun Zhu.
\newblock Perspectivenet: 3d object detection from a single rgb image via perspective points.
\newblock In \emph{Advances in Neural Information Processing Systems (NeurIPS)}, 2019.

\bibitem[Izadinia et~al.(2017)Izadinia, Shan, and Seitz]{izadinia2017im2cad}
Hamid Izadinia, Qi Shan, and Steven~M Seitz.
\newblock Im2cad.
\newblock In \emph{Conference on Computer Vision and Pattern Recognition (CVPR)}, 2017.

\bibitem[Jun and Nichol(2023)]{jun2023shape}
Heewoo Jun and Alex Nichol.
\newblock Shap-e: Generating conditional 3d implicit functions.
\newblock \emph{arXiv preprint arXiv:2305.02463}, 2023.

\bibitem[Kar et~al.(2017)Kar, H{\"a}ne, and Malik]{kar2017learning}
Abhishek Kar, Christian H{\"a}ne, and Jitendra Malik.
\newblock Learning a multi-view stereo machine.
\newblock In \emph{Advances in Neural Information Processing Systems (NeurIPS)}, 2017.

\bibitem[Kato et~al.(2020)Kato, Beker, Morariu, Ando, Matsuoka, Kehl, and Gaidon]{kato2020differentiable}
Hiroharu Kato, Deniz Beker, Mihai Morariu, Takahiro Ando, Toru Matsuoka, Wadim Kehl, and Adrien Gaidon.
\newblock Differentiable rendering: A survey.
\newblock \emph{arXiv preprint arXiv:2006.12057}, 2020.

\bibitem[Kingma and Ba(2014)]{kingma2014adam}
Diederik~P Kingma and Jimmy Ba.
\newblock Adam: A method for stochastic optimization.
\newblock In \emph{International Conference on Learning Representations (ICLR)}, 2014.

\bibitem[Knapitsch et~al.(2017)Knapitsch, Park, Zhou, and Koltun]{knapitsch2017tanks}
Arno Knapitsch, Jaesik Park, Qian-Yi Zhou, and Vladlen Koltun.
\newblock Tanks and temples: Benchmarking large-scale scene reconstruction.
\newblock \emph{ACM Transactions on Graphics (TOG)}, 36\penalty0 (4):\penalty0 1--13, 2017.

\bibitem[Lazova et~al.(2023)Lazova, Guzov, Olszewski, Tulyakov, and Pons-Moll]{lazova2023control}
Verica Lazova, Vladimir Guzov, Kyle Olszewski, Sergey Tulyakov, and Gerard Pons-Moll.
\newblock Control-nerf: Editable feature volumes for scene rendering and manipulation.
\newblock In \emph{Proceedings of Winter Conference on Applications of Computer Vision (WACV)}, 2023.

\bibitem[Lee et~al.(2009)Lee, Hebert, and Kanade]{lee2009geometric}
David~C Lee, Martial Hebert, and Takeo Kanade.
\newblock Geometric reasoning for single image structure recovery.
\newblock In \emph{Conference on Computer Vision and Pattern Recognition (CVPR)}, 2009.

\bibitem[Liu et~al.(2022)Liu, Zheng, Chen, Cui, and Han]{liu2022towards}
Haolin Liu, Yujian Zheng, Guanying Chen, Shuguang Cui, and Xiaoguang Han.
\newblock Towards high-fidelity single-view holistic reconstruction of indoor scenes.
\newblock In \emph{European Conference on Computer Vision (ECCV)}, 2022.

\bibitem[Liu et~al.(2020)Liu, Gu, Zaw~Lin, Chua, and Theobalt]{liu2020neural}
Lingjie Liu, Jiatao Gu, Kyaw Zaw~Lin, Tat-Seng Chua, and Christian Theobalt.
\newblock Neural sparse voxel fields.
\newblock In \emph{Advances in Neural Information Processing Systems (NeurIPS)}, 2020.

\bibitem[Liu et~al.(2023)Liu, Wu, Hoorick, Tokmakov, Zakharov, and Vondrick]{liu2023zero1to3}
Ruoshi Liu, Rundi Wu, Basile~Van Hoorick, Pavel Tokmakov, Sergey Zakharov, and Carl Vondrick.
\newblock Zero-1-to-3: Zero-shot one image to 3d object.
\newblock In \emph{International Conference on Computer Vision (ICCV)}, 2023.

\bibitem[Long et~al.(2023)Long, Lin, Liu, Liu, Wang, Theobalt, Komura, and Wang]{long2023neuraludf}
Xiaoxiao Long, Cheng Lin, Lingjie Liu, Yuan Liu, Peng Wang, Christian Theobalt, Taku Komura, and Wenping Wang.
\newblock Neuraludf: Learning unsigned distance fields for multi-view reconstruction of surfaces with arbitrary topologies.
\newblock In \emph{Conference on Computer Vision and Pattern Recognition (CVPR)}, 2023.

\bibitem[Lorensen and Cline(1987)]{lorensen1987marching}
William~E Lorensen and Harvey~E Cline.
\newblock Marching cubes: A high resolution 3d surface construction algorithm.
\newblock \emph{ACM Transactions on Graphics (TOG)}, 21\penalty0 (4):\penalty0 163--169, 1987.

\bibitem[Ma et~al.(2023)Ma, Yong, Zheng, Li, Liang, Zhu, and Huang]{ma2023sqa3d}
Xiaojian Ma, Silong Yong, Zilong Zheng, Qing Li, Yitao Liang, Song-Chun Zhu, and Siyuan Huang.
\newblock Sqa3d: Situated question answering in 3d scenes.
\newblock In \emph{International Conference on Learning Representations (ICLR)}, 2023.

\bibitem[Martin-Brualla et~al.(2021)Martin-Brualla, Radwan, Sajjadi, Barron, Dosovitskiy, and Duckworth]{martin2021nerf}
Ricardo Martin-Brualla, Noha Radwan, Mehdi~SM Sajjadi, Jonathan~T Barron, Alexey Dosovitskiy, and Daniel Duckworth.
\newblock Nerf in the wild: Neural radiance fields for unconstrained photo collections.
\newblock In \emph{Conference on Computer Vision and Pattern Recognition (CVPR)}, 2021.

\bibitem[Melas-Kyriazi et~al.(2023)Melas-Kyriazi, Laina, Rupprecht, and Vedaldi]{melas2023realfusion}
Luke Melas-Kyriazi, Iro Laina, Christian Rupprecht, and Andrea Vedaldi.
\newblock Realfusion: 360deg reconstruction of any object from a single image.
\newblock In \emph{Conference on Computer Vision and Pattern Recognition (CVPR)}, 2023.

\bibitem[Mescheder et~al.(2019)Mescheder, Oechsle, Niemeyer, Nowozin, and Geiger]{mescheder2019occupancy}
Lars Mescheder, Michael Oechsle, Michael Niemeyer, Sebastian Nowozin, and Andreas Geiger.
\newblock Occupancy networks: Learning 3d reconstruction in function space.
\newblock In \emph{Conference on Computer Vision and Pattern Recognition (CVPR)}, 2019.

\bibitem[Mildenhall et~al.(2021)Mildenhall, Srinivasan, Tancik, Barron, Ramamoorthi, and Ng]{mildenhall2021nerf}
Ben Mildenhall, Pratul~P Srinivasan, Matthew Tancik, Jonathan~T Barron, Ravi Ramamoorthi, and Ren Ng.
\newblock Nerf: Representing scenes as neural radiance fields for view synthesis.
\newblock \emph{Communications of the ACM}, 65\penalty0 (1):\penalty0 99--106, 2021.

\bibitem[Mitra et~al.(2018)Mitra, Kim, Yumer, Hueting, Carr, and Reddy]{mitra2018seethrough}
Niloy~J Mitra, Vladimir Kim, Ersin Yumer, Moos Hueting, Nathan Carr, and Pradyumna Reddy.
\newblock Seethrough: Finding objects in heavily occluded indoor scene images.
\newblock In \emph{International Conference on 3D Vision (3DV)}, 2018.

\bibitem[Newcombe et~al.(2011)Newcombe, Lovegrove, and Davison]{newcombe2011dtam}
Richard~A Newcombe, Steven~J Lovegrove, and Andrew~J Davison.
\newblock Dtam: Dense tracking and mapping in real-time.
\newblock In \emph{International Conference on Computer Vision (ICCV)}, 2011.

\bibitem[Nie et~al.(2020)Nie, Han, Guo, Zheng, Chang, and Zhang]{nie2020total3d}
Yinyu Nie, Xiaoguang Han, Shihui Guo, Yujian Zheng, Jian Chang, and Jian~Jun Zhang.
\newblock Total3dunderstanding: Joint layout, object pose and mesh reconstruction for indoor scenes from a single image.
\newblock In \emph{Conference on Computer Vision and Pattern Recognition (CVPR)}, 2020.

\bibitem[Niemeyer and Geiger(2021)]{niemeyer2021giraffe}
Michael Niemeyer and Andreas Geiger.
\newblock Giraffe: Representing scenes as compositional generative neural feature fields.
\newblock In \emph{Conference on Computer Vision and Pattern Recognition (CVPR)}, 2021.

\bibitem[Niemeyer et~al.(2019)Niemeyer, Mescheder, Oechsle, and Geiger]{niemeyer2019occupancy}
Michael Niemeyer, Lars Mescheder, Michael Oechsle, and Andreas Geiger.
\newblock Occupancy flow: 4d reconstruction by learning particle dynamics.
\newblock In \emph{International Conference on Computer Vision (ICCV)}, 2019.

\bibitem[Niemeyer et~al.(2020)Niemeyer, Mescheder, Oechsle, and Geiger]{niemeyer2020differentiable}
Michael Niemeyer, Lars Mescheder, Michael Oechsle, and Andreas Geiger.
\newblock Differentiable volumetric rendering: Learning implicit 3d representations without 3d supervision.
\newblock In \emph{Conference on Computer Vision and Pattern Recognition (CVPR)}, 2020.

\bibitem[Oechsle et~al.(2021)Oechsle, Peng, and Geiger]{oechsle2021unisurf}
Michael Oechsle, Songyou Peng, and Andreas Geiger.
\newblock Unisurf: Unifying neural implicit surfaces and radiance fields for multi-view reconstruction.
\newblock In \emph{International Conference on Computer Vision (ICCV)}, 2021.

\bibitem[Park et~al.(2019)Park, Florence, Straub, Newcombe, and Lovegrove]{park2019deepsdf}
Jeong~Joon Park, Peter Florence, Julian Straub, Richard Newcombe, and Steven Lovegrove.
\newblock Deepsdf: Learning continuous signed distance functions for shape representation.
\newblock In \emph{Conference on Computer Vision and Pattern Recognition (CVPR)}, 2019.

\bibitem[Paszke et~al.(2019)Paszke, Gross, Massa, Lerer, Bradbury, Chanan, Killeen, Lin, Gimelshein, Antiga, et~al.]{paszke2019pytorch}
Adam Paszke, Sam Gross, Francisco Massa, Adam Lerer, James Bradbury, Gregory Chanan, Trevor Killeen, Zeming Lin, Natalia Gimelshein, Luca Antiga, et~al.
\newblock Pytorch: An imperative style, high-performance deep learning library.
\newblock In \emph{Advances in Neural Information Processing Systems (NeurIPS)}, 2019.

\bibitem[Pfaff et~al.(2021)Pfaff, Fortunato, Sanchez-Gonzalez, and Battaglia]{pfaff2020learning}
Tobias Pfaff, Meire Fortunato, Alvaro Sanchez-Gonzalez, and Peter~W Battaglia.
\newblock Learning mesh-based simulation with graph networks.
\newblock In \emph{International Conference on Learning Representations (ICLR)}, 2021.

\bibitem[Poole et~al.(2023)Poole, Jain, Barron, and Mildenhall]{poole2022dreamfusion}
Ben Poole, Ajay Jain, Jonathan~T. Barron, and Ben Mildenhall.
\newblock Dreamfusion: Text-to-3d using 2d diffusion.
\newblock In \emph{International Conference on Learning Representations (ICLR)}, 2023.

\bibitem[Qi et~al.(2017)Qi, Su, Mo, and Guibas]{qi2017pointnet}
Charles~R Qi, Hao Su, Kaichun Mo, and Leonidas~J Guibas.
\newblock Pointnet: Deep learning on point sets for 3d classification and segmentation.
\newblock In \emph{Conference on Computer Vision and Pattern Recognition (CVPR)}, 2017.

\bibitem[Ranftl et~al.(2021)Ranftl, Bochkovskiy, and Koltun]{ranftl2021vision}
Ren{\'e} Ranftl, Alexey Bochkovskiy, and Vladlen Koltun.
\newblock Vision transformers for dense prediction.
\newblock In \emph{International Conference on Computer Vision (ICCV)}, 2021.

\bibitem[Saito et~al.(2019)Saito, Huang, Natsume, Morishima, Kanazawa, and Li]{saito2019pifu}
Shunsuke Saito, Zeng Huang, Ryota Natsume, Shigeo Morishima, Angjoo Kanazawa, and Hao Li.
\newblock Pifu: Pixel-aligned implicit function for high-resolution clothed human digitization.
\newblock In \emph{International Conference on Computer Vision (ICCV)}, 2019.

\bibitem[Shen et~al.(2023)Shen, Yang, and Wang]{shen2023anything3d}
Qiuhong Shen, Xingyi Yang, and Xinchao Wang.
\newblock Anything-3d: Towards single-view anything reconstruction in the wild.
\newblock \emph{arXiv preprint arXiv:2304.10261}, 2023.

\bibitem[Siddiqui et~al.(2022)Siddiqui, Thies, Ma, Shan, Nie{\ss}ner, and Dai]{siddiqui2022texturify}
Yawar Siddiqui, Justus Thies, Fangchang Ma, Qi Shan, Matthias Nie{\ss}ner, and Angela Dai.
\newblock Texturify: Generating textures on 3d shape surfaces.
\newblock In \emph{European Conference on Computer Vision (ECCV)}, 2022.

\bibitem[Song et~al.(2015)Song, Lichtenberg, and Xiao]{song2015sun}
Shuran Song, Samuel~P Lichtenberg, and Jianxiong Xiao.
\newblock Sun rgb-d: A rgb-d scene understanding benchmark suite.
\newblock In \emph{Conference on Computer Vision and Pattern Recognition (CVPR)}, 2015.

\bibitem[Sun et~al.(2018)Sun, Wu, Zhang, Zhang, Zhang, Xue, Tenenbaum, and Freeman]{sun2018pix3d}
Xingyuan Sun, Jiajun Wu, Xiuming Zhang, Zhoutong Zhang, Chengkai Zhang, Tianfan Xue, Joshua~B Tenenbaum, and William~T Freeman.
\newblock Pix3d: Dataset and methods for single-image 3d shape modeling.
\newblock In \emph{Conference on Computer Vision and Pattern Recognition (CVPR)}, 2018.

\bibitem[Tang(2022)]{tang2022stable}
Jiaxiang Tang.
\newblock Stable-dreamfusion: Text-to-3d with stable-diffusion, 2022.

\bibitem[Tang et~al.(2023)Tang, Wang, Zhang, Zhang, Yi, Ma, and Chen]{tang2023make}
Junshu Tang, Tengfei Wang, Bo Zhang, Ting Zhang, Ran Yi, Lizhuang Ma, and Dong Chen.
\newblock Make-it-3d: High-fidelity 3d creation from a single image with diffusion prior.
\newblock In \emph{International Conference on Computer Vision (ICCV)}, 2023.

\bibitem[Tewari et~al.(2020)Tewari, Fried, Thies, Sitzmann, Lombardi, Sunkavalli, Martin-Brualla, Simon, Saragih, Niessner, et~al.]{tewari2020state}
Ayush Tewari, Ohad Fried, Justus Thies, Vincent Sitzmann, Stephen Lombardi, Kalyan Sunkavalli, Ricardo Martin-Brualla, Tomas Simon, Jason Saragih, Matthias Niessner, et~al.
\newblock State of the art on neural rendering.
\newblock In \emph{Computer Graphics Forum}, 2020.

\bibitem[Wang et~al.(2023)Wang, Du, Li, Yeh, and Shakhnarovich]{wang2023score}
Haochen Wang, Xiaodan Du, Jiahao Li, Raymond~A Yeh, and Greg Shakhnarovich.
\newblock Score jacobian chaining: Lifting pretrained 2d diffusion models for 3d generation.
\newblock In \emph{Conference on Computer Vision and Pattern Recognition (CVPR)}, 2023.

\bibitem[Wang et~al.(2022)Wang, Wang, Long, Theobalt, Komura, Liu, and Wang]{wang2022neuris}
Jiepeng Wang, Peng Wang, Xiaoxiao Long, Christian Theobalt, Taku Komura, Lingjie Liu, and Wenping Wang.
\newblock Neuris: Neural reconstruction of indoor scenes using normal priors.
\newblock In \emph{European Conference on Computer Vision (ECCV)}, 2022.

\bibitem[Wang et~al.(2018)Wang, Zhang, Li, Fu, Liu, and Jiang]{wang2018pixel2mesh}
Nanyang Wang, Yinda Zhang, Zhuwen Li, Yanwei Fu, Wei Liu, and Yu-Gang Jiang.
\newblock Pixel2mesh: Generating 3d mesh models from single rgb images.
\newblock In \emph{European Conference on Computer Vision (ECCV)}, 2018.

\bibitem[Wang et~al.(2021)Wang, Liu, Liu, Theobalt, Komura, and Wang]{wang2021neus}
Peng Wang, Lingjie Liu, Yuan Liu, Christian Theobalt, Taku Komura, and Wenping Wang.
\newblock Neus: Learning neural implicit surfaces by volume rendering for multi-view reconstruction.
\newblock In \emph{Advances in Neural Information Processing Systems (NeurIPS)}, 2021.

\bibitem[Wu et~al.(2016)Wu, Zhang, Xue, Freeman, and Tenenbaum]{wu2016learning}
Jiajun Wu, Chengkai Zhang, Tianfan Xue, Bill Freeman, and Josh Tenenbaum.
\newblock Learning a probabilistic latent space of object shapes via 3d generative-adversarial modeling.
\newblock In \emph{Advances in Neural Information Processing Systems (NeurIPS)}, 2016.

\bibitem[Wu et~al.(2022)Wu, Liu, Chen, Li, Zheng, Cai, and Zheng]{wu2022object}
Qianyi Wu, Xian Liu, Yuedong Chen, Kejie Li, Chuanxia Zheng, Jianfei Cai, and Jianmin Zheng.
\newblock Object-compositional neural implicit surfaces.
\newblock In \emph{European Conference on Computer Vision (ECCV)}, 2022.

\bibitem[Xu et~al.(2019)Xu, Wang, Ceylan, Mech, and Neumann]{xu2019disn}
Qiangeng Xu, Weiyue Wang, Duygu Ceylan, Radomir Mech, and Ulrich Neumann.
\newblock Disn: Deep implicit surface network for high-quality single-view 3d reconstruction.
\newblock In \emph{Advances in Neural Information Processing Systems (NeurIPS)}, 2019.

\bibitem[Yang et~al.(2021)Yang, Zhang, Xu, Li, Zhou, Bao, Zhang, and Cui]{yang2021learning}
Bangbang Yang, Yinda Zhang, Yinghao Xu, Yijin Li, Han Zhou, Hujun Bao, Guofeng Zhang, and Zhaopeng Cui.
\newblock Learning object-compositional neural radiance field for editable scene rendering.
\newblock In \emph{International Conference on Computer Vision (ICCV)}, 2021.

\bibitem[Yariv et~al.(2020)Yariv, Kasten, Moran, Galun, Atzmon, Ronen, and Lipman]{yariv2020multiview}
Lior Yariv, Yoni Kasten, Dror Moran, Meirav Galun, Matan Atzmon, Basri Ronen, and Yaron Lipman.
\newblock Multiview neural surface reconstruction by disentangling geometry and appearance.
\newblock In \emph{Advances in Neural Information Processing Systems (NeurIPS)}, 2020.

\bibitem[Yariv et~al.(2021)Yariv, Gu, Kasten, and Lipman]{yariv2021volume}
Lior Yariv, Jiatao Gu, Yoni Kasten, and Yaron Lipman.
\newblock Volume rendering of neural implicit surfaces.
\newblock In \emph{Advances in Neural Information Processing Systems (NeurIPS)}, 2021.

\bibitem[Yu et~al.(2021)Yu, Ye, Tancik, and Kanazawa]{yu2021pixelnerf}
Alex Yu, Vickie Ye, Matthew Tancik, and Angjoo Kanazawa.
\newblock pixelnerf: Neural radiance fields from one or few images.
\newblock In \emph{Conference on Computer Vision and Pattern Recognition (CVPR)}, 2021.

\bibitem[Yu et~al.(2022)Yu, Peng, Niemeyer, Sattler, and Geiger]{yu2022monosdf}
Zehao Yu, Songyou Peng, Michael Niemeyer, Torsten Sattler, and Andreas Geiger.
\newblock Monosdf: Exploring monocular geometric cues for neural implicit surface reconstruction.
\newblock In \emph{Advances in Neural Information Processing Systems (NeurIPS)}, 2022.

\bibitem[Yuan et~al.(2022)Yuan, Sun, Lai, Ma, Jia, and Gao]{yuan2022nerf}
Yu-Jie Yuan, Yang-Tian Sun, Yu-Kun Lai, Yuewen Ma, Rongfei Jia, and Lin Gao.
\newblock Nerf-editing: geometry editing of neural radiance fields.
\newblock In \emph{Conference on Computer Vision and Pattern Recognition (CVPR)}, 2022.

\bibitem[Zamir et~al.(2020)Zamir, Sax, Cheerla, Suri, Cao, Malik, and Guibas]{zamir2020robust}
Amir~R Zamir, Alexander Sax, Nikhil Cheerla, Rohan Suri, Zhangjie Cao, Jitendra Malik, and Leonidas~J Guibas.
\newblock Robust learning through cross-task consistency.
\newblock In \emph{Conference on Computer Vision and Pattern Recognition (CVPR)}, 2020.

\bibitem[Zhang et~al.(2021)Zhang, Cui, Zhang, Zeng, Pollefeys, and Liu]{zhang2021holistic}
Cheng Zhang, Zhaopeng Cui, Yinda Zhang, Bing Zeng, Marc Pollefeys, and Shuaicheng Liu.
\newblock Holistic 3d scene understanding from a single image with implicit representation.
\newblock In \emph{Conference on Computer Vision and Pattern Recognition (CVPR)}, 2021.

\bibitem[Zhao et~al.(2022)Zhao, Cai, Zhang, Sheng, Xu, Zheng, Zhao, Wang, and Fan]{zhao20223dgqa}
Lichen Zhao, Daigang Cai, Jing Zhang, Lu Sheng, Dong Xu, Rui Zheng, Yinjie Zhao, Lipeng Wang, and Xibo Fan.
\newblock Towards explainable 3d grounded visual question answering: A new benchmark and strong baseline.
\newblock \emph{IEEE Transactions on Circuits and Systems for Video Technology}, 2022.

\end{thebibliography}
}

\clearpage

\renewcommand\thefigure{S\arabic{figure}}
\setcounter{figure}{0}
\renewcommand\thetable{S\arabic{table}}
\setcounter{table}{0}
\renewcommand\theequation{S\arabic{equation}}
\setcounter{equation}{0}
\renewcommand{\thesection}{S\arabic{section}}
\setcounter{section}{0}
\setcounter{footnote}{0}
\setcounter{page}{1}

\maketitlesupplementary

In \cref{supp:dataset}, we present how we prepare the training data on the \threedfront~\cite{fu20213d} and \pixthreed~\cite{sun2018pix3d} datasets.
In \cref{supp:tech_detail}, we report more implementation details, including model architecture, learning curriculum, and training strategies.
\cref{supp:experiment details} presents more experimental details, evaluation metrics, results, and failure case discussion.
For a more comprehensive view of the qualitative results, we recommend referring to the supplementary video with detailed visualizations and animations.

\section{Training data preparation}\label{supp:dataset}

\subsection{Datasets and splits}

\threedfront~\cite{fu20213d} contains synthetic and professionally-designed indoor scenes populated by high-quality textured 3D models from \threedfuture~\cite{fu20213d}.
Following Liu \etal~\cite{liu2022towards}, we use about 20K scene images for training and validation and report quantitative evaluation on 2000 images in 8 different categories.
\pixthreed~\cite{sun2018pix3d} provides real-world images along with corresponding 3D furniture models that are aligned with the images in 9 object categories. In the splits from Nie \etal~\cite{nie2020total3d} and ~\cite{zhang2021holistic}, there is a significant overlap of objects between the training and testing split. Consequently, we follow Liu \etal ~\cite{liu2022towards} to employ a split without overlapping objects.

\subsection{Fixing CAD models}

One major issue with \threedfront~\cite{fu20213d} dataset is that the majority of the CAD models are not watertight, which hinders the learning of \ac{sdf} as the neural implicit surface representations in our framework.
To mitigate this issue, we first utilize the automatic remesh method following Liu \etal~\cite{liu2022towards} to transform the non-watertight models into watertight ones with a more evenly-distributed topology. Additionally, we have noticed that some models, primarily beds and sofas, lack a back or underside, confusing the model when learning 3D object priors. Consequently, we have manually fixed a total of 1734 models to resolve this issue.

\subsection{SDF supervision}

In our framework, we employ supervision from both explicit 3D shapes and volume rendering of color, depth, and normal images. 
The direct 3D supervision focuses on minimizing the differences between the predicted and actual \ac{sdf} values of the sampled points. 
To generate \gt \ac{sdf} values for a given CAD model, we voxelize its 3D bounding box in 
a normalized coordinate system with a resolution of 64.
The \ac{sdf} value for each voxel grid center is calculated as the distance to the mesh surface.
The \ac{sdf} value is positive if the point is outside the surface and negative if inside.
During training, we obtain the \gt \ac{sdf} value for the query points using trilinear interpolation.

\subsection{Monocular cues rendering}\label{supp:render_method}

To further alleviate the ambiguities in recovering 3D shapes from single-view inputs, we follow Yu \etal~\cite{yu2022monosdf} to exploit monocular depth, normal, and segmentation cues to facilitate the training process.
However, since these images are not available in the \threedfront~\cite{fu20213d} dataset, we render them using the 3D scans of the scene, 3D CAD models of the objects, and the camera's intrinsic and extrinsic parameters provided in the dataset.
The \pixthreed~\cite{sun2018pix3d} dataset offers instance segmentation but lacks depth and normal images. 
Since rendering is impossible, we utilize the estimated depth and normal maps as the pseudo-\gt from \sota estimator~\cite{eftekhar2021omnidata}.
Note that the depth, normal, and segmentation information is solely used during the training stage to guide the model's learning process, and none is required during the inference stage. This ensures that our model remains flexible and applicable to various scenarios.

\section{Technical details}\label{supp:tech_detail}

\subsection{Model architecture}

For the implicit network, we use an $8$-layer MLP with hidden dimension $256$. We implement the rendering network with a $2$-layer MLP with hidden dimension 256. We use Softplus activation for the implicit network and Sigmoid activation for the rendering network. We use a ResNet34 backbone pre-trained on ImageNet as the image encoder following Yu \etal~\cite{yu2021pixelnerf}. We use positional encoding $\gamma$ from \ac{nerf}~\cite{mildenhall2021nerf} for the spatial coordinates, with $L=6$ exponentially increasing frequencies:
\begin{equation}
    \begin{aligned}
        \gamma(x) = (&\sin(2^0\omega \textbf{x}), \cos(2^0\omega\\
        &\sin(2^1\omega \textbf{x}), \cos(2^1\omega\textbf{x}), \\&\ldots, \\
        &\sin(2^{L-1}\omega \textbf{x}), \cos(2^{L-1}\omega \textbf{x}))
    \end{aligned}
\end{equation}

\subsection{Learning curriculum}

As discussed in the paper, we propose a two-stage learning curriculum to effectively employ supervision from both 3D shapes and volume rendering. 
In \stageone, the loss weight is set to be $1$ for the 3D supervision $\mathcal{L}_{3D}$ and $0$ for the rest of the losses. In \stagetwo, we linearly increase the loss weights for color, depth, and normal supervision while maintaining a constant weight for the 3D supervision loss. We utilize the $\mathcal{L}_{3D}$ loss curve when training with 3D supervision only to determine the suitable value for $\lambda_0$. Specifically, once the training approaches convergence according to the \ac{sdf} loss curve, we identify the epoch at this point as the suitable $\lambda_0$ value. In our paper, we choose $\lambda_0 = 150$. Slight deviations below or above this threshold have minimal impact. However, significantly reducing the value, such as $\lambda_0 = 0$ or $\lambda_0 = 70$, results in substantial degradation of performance because early injection of 2D supervision may affect the 3D shape learning due to the shape-appearance ambiguity. More discussion can be referred to the ablation experiments in \cref{sec:exp_obj_recon}.

\subsection{Training strategy}

During training, the image, depth, and normal images have the same resolution of $484\times648$. The implicit network takes 3D points as input in canonical coordinate system to ease the learning of reconstructing indoor objects with implicit representations. The volume rendering is performed along the sampled points in the camera coordinates to calculate the color, depth, and normal values for all the pixels in a minibatch. We sample 64 rays per iteration and apply the error-bounded sampling strategy introduced by Yariv \etal~\cite{yariv2021volume}. We additionally apply 3D supervision to another 30,000 points uniformly sampled near the \gt surfaces (set $\mathcal{X}$). Our model is trained for 400 epochs on 4 NVIDIA-A100 GPUs with a batch size of 96. We implement our method in PyTorch~\cite{paszke2019pytorch} and use the Adam optimizer~\cite{kingma2014adam}. 
The learning rate is initialized as 1e-3 and decays by a factor of 0.2 in the $330^{th}$ and $370^{th}$ epochs.

\section{More experiment details and results}\label{supp:experiment details}

\subsection{Evaluation metrics}

\paragraph{3D object reconstruction}

Following Wang \etal~\cite{wang2018pixel2mesh} and Mescheder \etal\cite{mescheder2019occupancy}, we adopt \acf{cd}, \fscore and \normalconsist as the metrics to evaluate 3D object reconstruction.
Following prior work~\cite{liu2022towards}, we proportionally scale the longest edge of reconstructed objects to $2m$ to calculate \ac{cd} and \fscore.
After mesh alignment with Iterative Closest Point (ICP)~\cite{besl1992method}, we uniformly sample points from our prediction and \gt.
\ac{cd} is calculated by summing the squared distances between the nearest neighbor correspondences of two point clouds after mesh alignment. The values of \ac{cd} are reported in units of $10^{-3}$.
We calculate precision and recall by checking the percentage of points in prediction or \gt that can find the nearest neighbor from the other within a threshold of $2mm$. 
\fscore~\cite{knapitsch2017tanks} is the harmonic mean of precision and recall on in prediction and \gt. 
Finally, to measure how well the methods can capture higher-order information, the normal consistency score is computed as the mean absolute dot product of the normals in one mesh and the normals at the corresponding nearest neighbors in the other mesh after alignment.

\paragraph{Depth and normal estimation}

We adopt the L1 error for depth estimation and utilize both L1 and Angular errors for normal estimation following Eftekhar \etal~\cite{eftekhar2021omnidata}. 
Since the baseline methods~\cite{eftekhar2021omnidata,zamir2020robust} estimate relative depth values rather than absolute ones, we first align the estimated depth values with the \gt values to the range of $[0, 1]$ using the approach outlined in Eftekhar \etal~\cite{eftekhar2021omnidata}. After the alignment, we compute the L1 error to quantify the discrepancy. For normal estimation, we normalize both the estimated and \gt values to unit vectors and then compute the L1 error and the angle error for evaluation.

\subsection{Indoor object reconstruction}

\subsubsection{Experiments on \threedfront and \pixthreed}

In this subsection, we provide more details about our reproduced results and more qualitative results. \cref{tab:original_front3d,tab:original_pix3d} list the quantitative outcomes on \threedfront~\cite{fu20213d} and \pixthreed~\cite{sun2018pix3d} datasets documented by Liu \etal~\cite{liu2022towards} in the original paper. The ``NC'' (normal consistency) columns in these tables are empty, as Liu \etal~\cite{liu2022towards} did not measure normal consistency in their paper. \cref{tab:reproduce_front3d,tab:reproduce_pix3d} present our reproduced results, which are comparable with the results in \cref{tab:original_front3d,tab:original_pix3d}.
We additionally provide more qualitative results, \cref{fig:supp_qual_front3d} on \threedfront and \cref{fig:supp_qual_pix3d} on \pixthreed. As can be seen from these figures, our model can learn finer and smoother surfaces with high-fidelity textures.

\begin{table}[b!]
    \centering
    \small
    \setlength{\tabcolsep}{3pt}
    \caption{\textbf{Object reconstruction on the \threedfront~\cite{fu20213d} dataset.} Our model achieves the best performance on mean \acs{cd} and \fscore, as well as the best \acs{nc} on all object categories, outperforming MGN~\cite{nie2020total3d}, LIEN~\cite{zhang2021holistic}, and InstPIFu~\cite{liu2022towards}.}
    \resizebox{\linewidth}{!}{%
        \begin{tabular}{ccccccccccc}
            \toprule
            \multicolumn{2}{c}{\textbf{Category}} & bed & chair & sofa & table & desk & nightstand & cabinet & bookshelf & mean \\
            \midrule
            \multirow{4}{*}{\textbf{\acs{cd} $\downarrow$}} & MGN  & 15.48  & 11.67  & 8.72  & 20.90  & 17.59  & 17.11  & 13.13  & 10.21  & 14.07  \\
            & LIEN & 16.81  & 41.40  & 9.51  & 35.65  & 26.63  & 16.78  & 7.44  & 11.70  & 28.52  \\
            & InstPIFu & 18.17  & 14.06  & 7.66  & 23.25  & 33.33  & \textbf{11.73}  & \textbf{6.04}  & 8.03  & 14.46  \\
            & Ours & \textbf{4.96}  & \textbf{10.52}  & \textbf{4.53}  & \textbf{16.12}  & \textbf{25.86}  & 17.90  & 6.79  & \textbf{3.89}  & \textbf{10.45}  \\
            \midrule
            \multirow{4}{*}{\textbf{$\fscore\uparrow$}} & MGN & 46.81  & 57.49  & 64.61  & 49.80  & 46.82  & 47.91  & 54.18  & 54.55  & 55.64  \\
            & LIEN & 44.28  & 31.61  & 61.40  & 43.22  & 37.04  & 50.76  & 69.21  & 55.33  & 45.63  \\
            & InstPIFu & 47.85  & 59.08  & 67.60  & 56.43  & \textbf{48.49}  & 57.14  & \textbf{73.32}  & 66.13  & 61.32  \\
            & Ours & \textbf{76.34}  & \textbf{69.17}  & \textbf{80.06}  & \textbf{67.29}  & 47.12  & \textbf{58.48}  & 70.45  & \textbf{85.93}  & \textbf{71.36}  \\
            \midrule
            \multirow{4}{*}{\textbf{$\acs{nc}\uparrow$}}     & MGN & -  & -  & -  & -  & -  & -  & -  & -  & -  \\
            & LIEN & -  & -  & -  & -  & -  & -  & -  & -  & -  \\
            & InstPIFu & -  & -  & -  & -  & -  & -  & -  & -  & -  \\
            & Ours & \textbf{0.896}  & \textbf{0.833}  & \textbf{0.894}  & \textbf{0.838}  & \textbf{0.764}  & \textbf{0.897}  & \textbf{0.856}  & \textbf{0.862}  & \textbf{0.854}  \\ 
            \bottomrule
        \end{tabular}%
    }%
    \label{tab:original_front3d}
\end{table}

\begin{table}[b!]
    \centering
    \small
    \setlength{\tabcolsep}{3pt}
    \caption{\textbf{Reproduced results on the \threedfront~\cite{fu20213d} dataset.} $\dagger$: Results reproduced from the official repository.}
    \resizebox{\linewidth}{!}{%
        \begin{tabular}{ccccccccccc}
            \toprule
            \multicolumn{2}{c}{\textbf{Category}} & bed & chair & sofa & table & desk & nightstand & cabinet & bookshelf & mean \\
            \midrule
            \multirow{4}{*}{\textbf{\acs{cd} $\downarrow$}} & MGN$^\dagger$  & 5.95  & 14.31  & 6.01  & 24.21  & 38.07  & 20.27  & 12.49  & 11.42  & 14.97  \\
            & LIEN$^\dagger$ & 4.58  & 11.85  & 7.21  & 30.40  & 42.52  & 20.84  & 14.81  & 18.13  & 16.33  \\
            & InstPIFu$^\dagger$ & 7.68  & 13.79  & 6.78  & 21.56  & 31.32  & \textbf{13.14}  & \textbf{5.94}  & 13.79  & 13.63  \\
            & Ours & \textbf{4.96}  & \textbf{10.52}  & \textbf{4.53}  & \textbf{16.12}  & \textbf{25.86}  & 17.90  & 6.79  & \textbf{3.89}  & \textbf{10.45}  \\
            \midrule
            \multirow{4}{*}{\textbf{$\fscore\uparrow$}} & MGN$^\dagger$ & 65.47  & 52.83  & 68.98  & 54.04  & 42.9  & 45.01  & 52.01  & 55.26  & 57.19  \\
            & LIEN$^\dagger$ & 72.16  & 62.23  & 68.25  & 48.18  & 32.07  & 42.87  & 49.62  & 43.12  & 58.37  \\
            & InstPIFu$^\dagger$ & 62.28  & 66.31  & 69.65  & 58.73  & 40.49  & 57.52  & \textbf{76.59}  & 71.48  & 65.34  \\
            & Ours & \textbf{76.34}  & \textbf{69.17}  & \textbf{80.06}  & \textbf{67.29}  & \textbf{47.12}  & \textbf{58.48}  & 70.45  & \textbf{85.93}  & \textbf{71.36}  \\
            \midrule
            \multirow{4}{*}{\textbf{$\acs{nc}\uparrow$}}     & MGN$^\dagger$ & 0.829  & 0.758  & 0.819  & 0.785  & 0.711  & 0.833  & 0.802  & 0.719  & 0.787  \\
            & LIEN$^\dagger$ & 0.822  & 0.793  & 0.803  & 0.755  & 0.701  & 0.814  & 0.801  & 0.747  & 0.786  \\
            & InstPIFu$^\dagger$ & 0.799  & 0.782  & 0.846  & 0.804  & 0.708  & 0.844  & 0.841  & 0.790  & 0.810  \\
            & Ours & \textbf{0.896}  & \textbf{0.833}  & \textbf{0.894}  & \textbf{0.838}  & \textbf{0.764}  & \textbf{0.897}  & \textbf{0.856}  & \textbf{0.862}  & \textbf{0.854}  \\ 
            \bottomrule
        \end{tabular}%
    }%
    \label{tab:reproduce_front3d}
\end{table}

\begin{table}[ht!]
    \centering
    \small
    \setlength{\tabcolsep}{3pt}
    \caption{\textbf{Object Reconstruction on the \pixthreed~\cite{sun2018pix3d} dataset.} On the non-overlapped split~\cite{liu2022towards}, our model outperforms the state-of-the-art methods by significant margins. }
    \resizebox{\linewidth}{!}{%
        \begin{tabular}{lcccccccccccc}
            \toprule
            \multicolumn{2}{c}{\textbf{Category}} & bed & bookcase & chair & desk & sofa & table & tool & wardrobe & misc & mean \\
            \midrule
            \multirow{4}{*}{\textbf{\acs{cd} $\downarrow$}} & MGN & 22.91 & 33.61 & 56.47 & 33.95 & 9.27 & 81.19 & 94.70 & 10.43 & 137.50 & 44.32 \\
            & LIEN & 11.18 & 29.61 & 40.01 & 65.36 & 10.54 & 146.13 & 29.63 & 4.88 & 144.06 & 51.31 \\
            & InstPIFu & 10.90 & 7.55 & 32.44 & \textbf{22.09} & 8.13 & 45.82 & 10.29 & \textbf{1.29} & 47.31 & 24.65 \\
            & Ours & \textbf{6.31} & \textbf{7.21} & \textbf{26.23} & 28.63 & \textbf{5.68} & \textbf{43.87} & \textbf{8.29} & 2.07 & \textbf{35.03} & \textbf{21.79} \\
            \midrule
            \multirow{4}{*}{\textbf{$\fscore\uparrow$}} & MGN & 34.69 & 28.42 & 35.67 & 65.36 & 51.15 & 17.05 & 57.16 & 52.04 & 10.41 & 36.20 \\
            & LIEN & 37.13 & 15.51 & 25.70 & 26.01 & 49.71 & 21.16 & 5.85 & 59.46 & 11.04 & 31.45 \\
            & InstPIFu & 54.99 & 62.26 & 35.30 & \textbf{47.30} & 56.54 & 37.51 & 64.24 & \textbf{94.62} & 27.03 & 45.62 \\
            & Ours & \textbf{68.78} & \textbf{66.69} & \textbf{55.18} & 42.49 & \textbf{71.22} & \textbf{51.93} & \textbf{65.38} & 91.84 & \textbf{46.92} & \textbf{59.71} \\
            \midrule
            \multirow{4}{*}{\textbf{$\acs{nc}\uparrow$}} & MGN & - & - & - & - & - & - & - & - & - & - \\
            & LIEN & - & - & - & - & - & - & - & - & - & - \\
            & InstPIFu & - & - & - & - & - & - & - & - & - & - \\
            & Ours & \textbf{0.825} & \textbf{0.689} & \textbf{0.693} & \textbf{0.776} & \textbf{0.866} & \textbf{0.835} & \textbf{0.645} & \textbf{0.960} & \textbf{0.599} & \textbf{0.778} \\
            \bottomrule
        \end{tabular}%
    }%
    \label{tab:original_pix3d}
\end{table}

\begin{table}[ht!]
    \centering
    \small
    \setlength{\tabcolsep}{3pt}
    \caption{\textbf{Reproduced results on the \pixthreed~\cite{sun2018pix3d} dataset.} $\dagger$: Results reproduced from the official repository.}
    \resizebox{\linewidth}{!}{%
        \begin{tabular}{lcccccccccccc}
            \toprule
            \multicolumn{2}{c}{\textbf{Category}} & bed & bookcase & chair & desk & sofa & table & tool & wardrobe & misc & mean \\
            \midrule
            \multirow{4}{*}{\textbf{\acs{cd} $\downarrow$}} & MGN$^\dagger$ & 11.73 & 17.50 & 36.74 & 29.63 & 7.02 & 47.85 & 25.63 & 6.61 & 62.32 & 27.9 \\
            & LIEN$^\dagger$ & 16.31 & 26.05 & 33.06 & 39.84 & 7.34 & 76.97 & 27.84 & 4.66 & 126.88 & 33.66 \\
            & InstPIFu$^\dagger$ & 9.82 & 7.73 & 31.55 & \textbf{23.18} & 7.28 & 45.89 & 9.64 & \textbf{1.79} & 43.4 & 24.35 \\
            & Ours & \textbf{6.31} & \textbf{7.21} & \textbf{26.23} & 28.63 & \textbf{5.68} & \textbf{43.87} & \textbf{8.29} & 2.07 & \textbf{35.03} & \textbf{21.79} \\
            \midrule
            \multirow{4}{*}{\textbf{$\fscore\uparrow$}} & MGN$^\dagger$ & 51.51 & 40.85 & 49.42 & 44.06 & 59.48 & 45.02 & 45.52 & 64.76 & 26.23 & 50.55 \\
            & LIEN$^\dagger$ & 50.20 & 36.26 & 48.41 & 32.55 & 64.98 & 29.97 & 38.72 & 76.03 & 18.29 & 47.87 \\
            & InstPIFu$^\dagger$ & 57.02 & 61.45 & 38.06 & \textbf{45.98} & 62.76 & 37.26 & 63.50 & \textbf{93.94} & 40.14 & 48.09 \\
            & Ours & \textbf{68.78} & \textbf{66.69} & \textbf{55.18} & 42.49 & \textbf{71.22} & \textbf{51.93} & \textbf{65.38} & 91.84 & \textbf{46.92} & \textbf{59.71} \\
            \midrule
            \multirow{4}{*}{\textbf{$\acs{nc}\uparrow$}} & MGN$^\dagger$ & 0.737 & 0.592 & 0.525 & 0.633 & 0.756 & 0.794 & 0.531 & 0.809 & 0.563 & 0.659 \\
            & LIEN$^\dagger$ & 0.706 & 0.514 & 0.591 & 0.581 & 0.775 & 0.619 & 0.506 & 0.844 & 0.481 & 0.646 \\
            & InstPIFu$^\dagger$ & 0.782 & 0.646 & 0.547 & 0.758 & 0.753 & 0.796 & 0.639 & 0.951 & 0.580 & 0.683 \\
            & Ours & \textbf{0.825} & \textbf{0.689} & \textbf{0.693} & \textbf{0.776} & \textbf{0.866} & \textbf{0.835} & \textbf{0.645} & \textbf{0.960} & \textbf{0.599} & \textbf{0.778} \\
            \bottomrule
        \end{tabular}%
    }%
    \label{tab:reproduce_pix3d}
\end{table}

\subsubsection{Failure cases}

\begin{figure}[ht!]
	\centering
	\includegraphics[width=\linewidth]{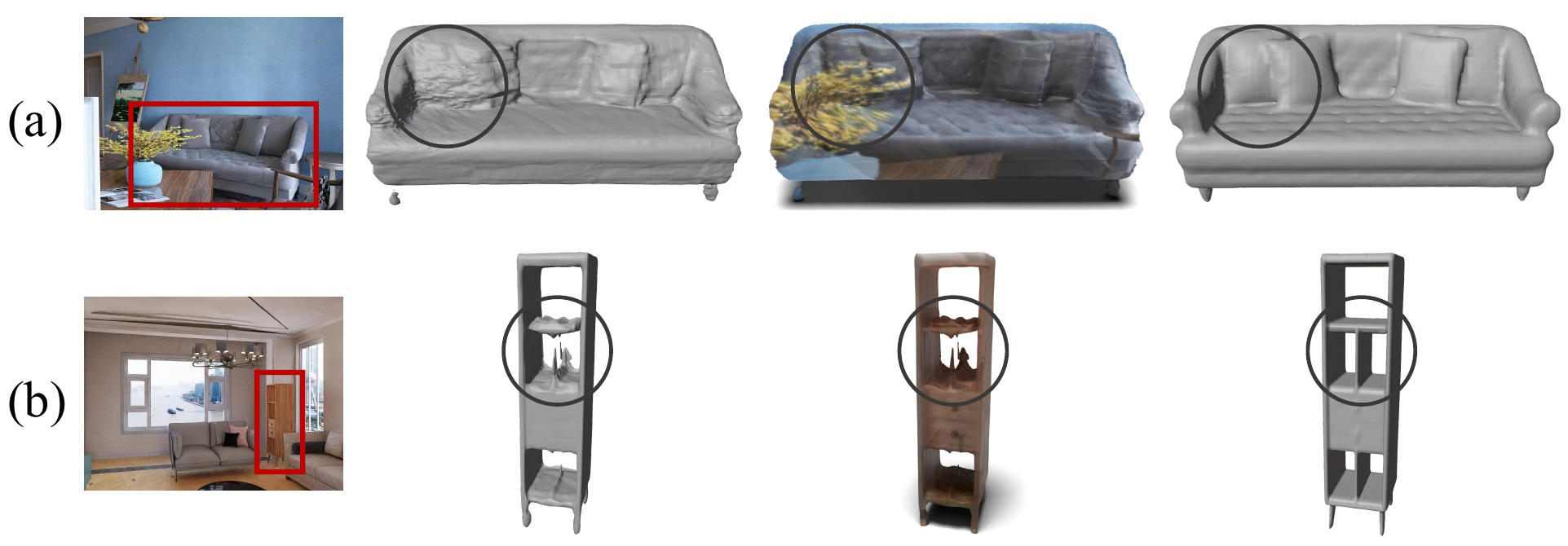}
        {\footnotesize \hspace*{0.8cm} Input \hspace{0.85cm} Ours$_{\text{shape}}$ \hspace{0.6cm}  Ours$_{\text{shape+texture}}$ \hspace{0.90cm} GT   \hfill}
	\caption{\textbf{Qualitative examples for failure cases.} Two representatives with (a) occlusions and (b) intricate geometry.}
	\label{fig:failure_cases}
\end{figure}

In this section, we present and diagnose some representative failure examples. \cref{fig:failure_cases}{\color{red}(a)} illustrates that occlusion between objects can lead to distortions in both shape and appearance recovery, particularly in the occluded areas. Prior work~\cite{liu2023zero1to3,jun2023shape} necessitates unoccluded object images as input to avoid this problem. One of the future directions is to involve the accurate reconstruction of object shapes with textures under heavy occlusions. The bookshelf depicted in \cref{fig:failure_cases}{\color{red}(b)} exemplifies the challenges our framework encounters when attempting to reconstruct intricate geometry. We hypothesize that this can be attributed to the limited representation power of \ac{sdf} as the implicit surface representations for thin surfaces. Specifically, it requires the model to recognize and reconstruct abrupt changes in the signed distance field within centimeters, transitioning from positive (outside) to negative (inside) and then back to positive again. To address such issues, we suggest future endeavors in integrating unsigned distance field~\cite{guillard2022meshudf,long2023neuraludf} with volume rendering to capture such intricate geometry and non-watertight meshes.

\subsubsection{Comparison with prior-guided models}

We randomly select 100 samples from the test split in \threedfront to perform a comparative assessment of the reconstruction performance between our framework and prior-guided models, \ie, Zero-1-to-3~\cite{liu2023zero1to3} and \shapee~\cite{jun2023shape}. In the original paper, Zero-1-to-3~\cite{liu2023zero1to3} utilized SJC~\cite{wang2023score} for the 3D reconstruction task, whereas we follow Tang \etal~\cite{tang2022stable} to employ DreamFusion~\cite{poole2022dreamfusion} to achieve enhanced results. In \shapee~\cite{jun2023shape}, the text prompts are required for reconstruction using the conditional generative model. We utilize the \gt object category as the designated text prompt to maintain fairness. It is noteworthy that both Zero-1-to-3~\cite{liu2023zero1to3} and \shapee~\cite{jun2023shape} necessitate background-free input images for the target objects. Consequently, we utilize the ground truth mask to segment the object as input. For our model, we use the original image as input.

\subsection{Rendering capability}

\subsubsection{Novel view synthesis}

\begin{figure*}[ht!]
	\centering
	\includegraphics[width=0.87\linewidth]{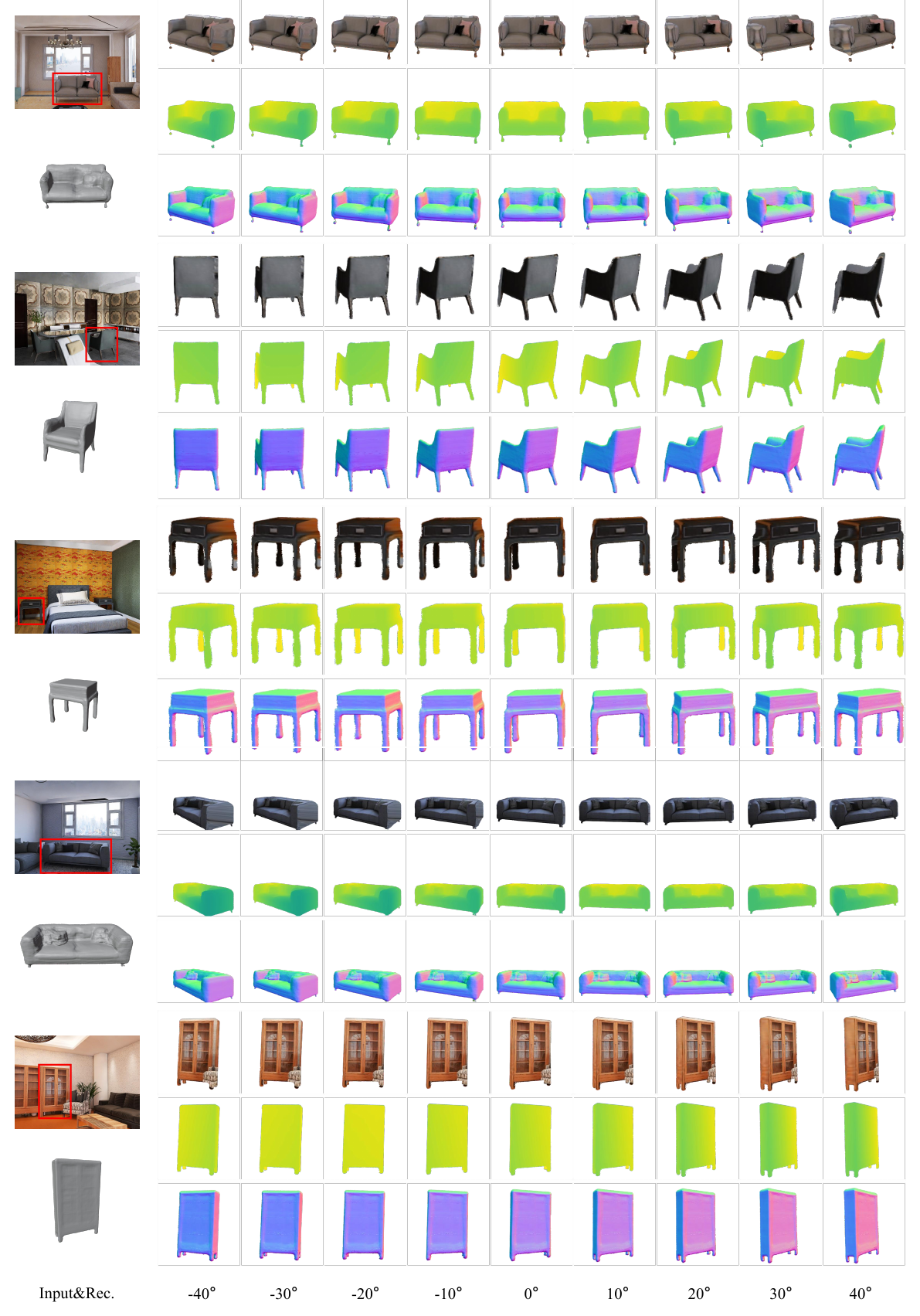}
	\caption{
	\textbf{More qualitative results for novel views synthesis.}
	}
	\label{fig:supp_novel_views}
\end{figure*}

To measure the capability for novel view synthesis, we compare our model with PixelNeRF~\cite{yu2021pixelnerf} on the category-agnostic model, which is first trained on the ShapeNet~\cite{shapenet2015} and then fine-tuned using the \threedfront~\cite{fu20213d}. 
Results in \cref{fig:pixelnerf} show that PixelNeRF struggles to render images outside the vicinity of the original viewpoints where our model is capable of generating meaningful renderings from novel views.
The main reason behind this is that our method employs 3D supervision, which helps the model learn better 3D object priors. PixelNeRF fails to acquire a meaningful 3D prior from the training data, especially when each image exists independently in \threedfront~\cite{fu20213d}, which is in stark contrast to the ShapeNet~\cite{shapenet2015} where images are presented in a sequence of related perspectives. \cref{fig:supp_novel_views} shows that our model not only generates high-quality new perspective images but also produces reasonable depth and normal maps by volume rendering.

\subsubsection{Depth and normal estimation}

\begin{figure}[h!]
	\centering
	\includegraphics[width=0.9\linewidth]{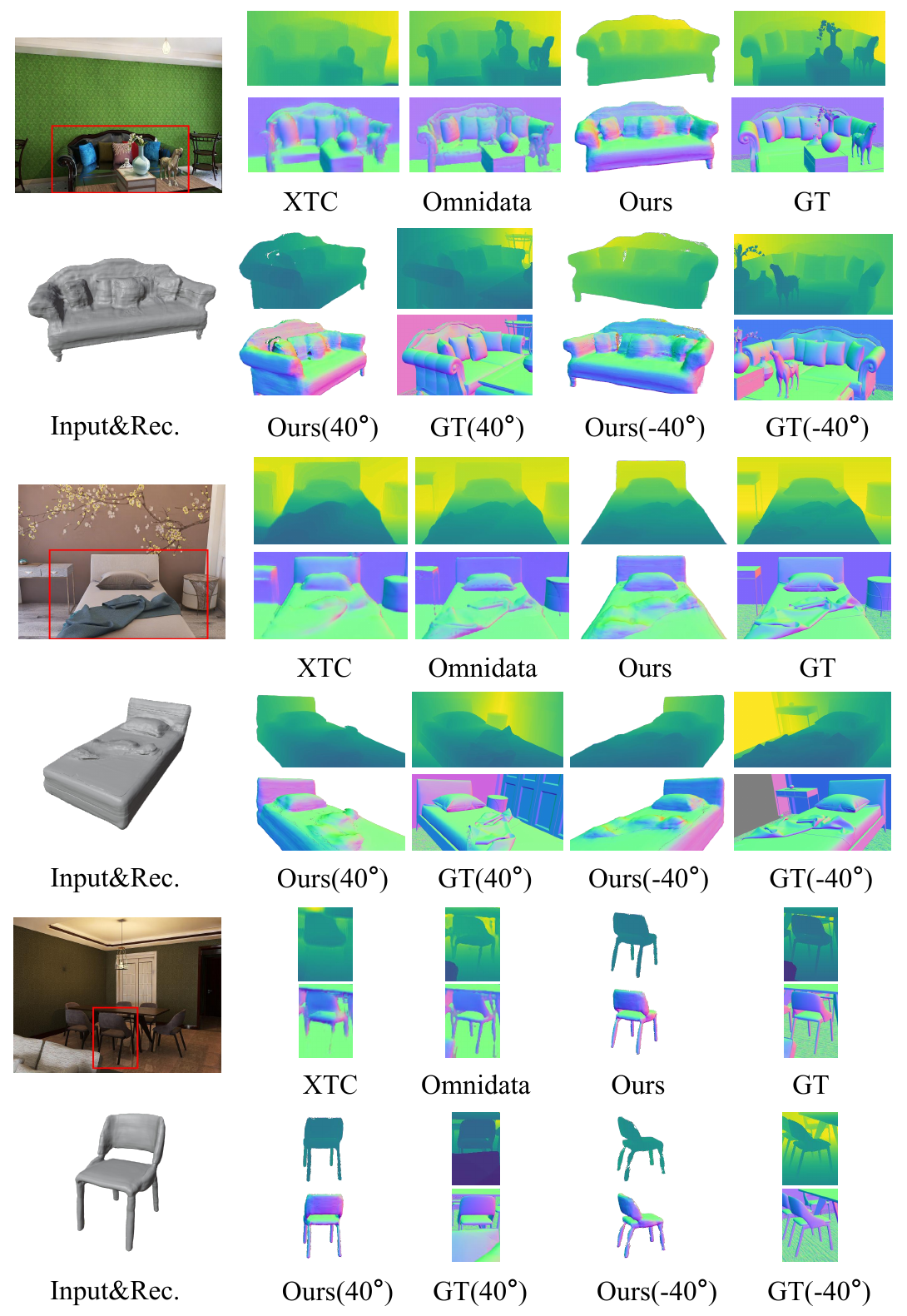}
	\caption{
	\textbf{Qualitative results for depth and normal estimation.} Our method produces results comparable with~\cite{zamir2020robust,eftekhar2021omnidata} on the input view, and can estimate depth and normal for novel views.
	}
	\label{fig:supp_dn}
\end{figure}

Our framework can serve as a proficient single-view depth and normal estimator, and to compare with zero-shot \sota methods~\cite{zamir2020robust,eftekhar2021omnidata}, we randomly select 200 samples from the test split in \threedfront. To assess our model's capability to generate depth and normal maps from novel views, we rotated the camera vertically left and right by 5$^{\circ}$, 10$^{\circ}$, 15$^{\circ}$, 20$^{\circ}$, 30$^{\circ}$, and 40$^{\circ}$. It is worth noting that only our method possesses the capability to estimate depth and normal from novel views.
\cref{tab:novel_depth_normal,fig:supp_dn} show that our model can consistently produce satisfactory outcomes, even when the viewing angle changes significantly.

\begin{table}[t!]
    \centering
    \small
    \caption{\textbf{Novel views depth and normal estimation.} We evaluate depth using L1 $\downarrow$ and normal using L1$\downarrow$ / Angular$^{\circ}$$\downarrow$ error.}
    \resizebox{\linewidth}{!}{%
        \begin{tabular}{lcccccc}
        \toprule
           Angle & Method & Depth(L1$\downarrow$) & Normal(L1$\downarrow$ / Angular$^{\circ}$$\downarrow$)\\
        \midrule
        \multirow{3}{*}{0$^{\circ}$} 
        & XTC~\cite{zamir2020robust} & 1.188        & 12.712 / 14.309\\
        & Omnidata~\cite{eftekhar2021omnidata} & 0.734        & 10.015 / 11.257\\
        & Ours                                 & 0.992        & 10.962 / 12.392\\
        \midrule
            5$^{\circ}$ & \multirow{6}{*}{Ours} & 1.0313 & 11.2132 / 12.631\\
            10$^{\circ}$ & & 1.0911 & 11.5866 / 13.1078\\
            15$^{\circ}$ & & 1.1796 & 12.0943 / 13.7095\\
            20$^{\circ}$ & & 1.2624 & 12.6315 / 14.2108\\
            30$^{\circ}$ & & 1.3934 & 13.4501 / 15.1012\\
            40$^{\circ}$ & & 1.4694 & 15.0034 / 16.6108\\
        \bottomrule
        \end{tabular}%
    }%
    \label{tab:novel_depth_normal}
\end{table}

\clearpage

\begin{figure*}[ht!]
    \centering
	\includegraphics[width=0.92\linewidth]{sup-main-front3d-v5}
        {\footnotesize \hspace*{1.4cm} Input \hspace{1.7cm}  MGN  \hspace{1.4cm} LIEN \hspace{1.4cm}  InstPIFu \hspace{1.1cm}  Ours$_{\text{shape}}$ \hspace{0.9cm} Ours$_{\text{shape+texture}}$ \hspace{1.0cm} GT \hfill }
	\caption{
	\textbf{More qualitative results from \threedfront~\cite{liu2022towards}}.
	}
	\label{fig:supp_qual_front3d}
\end{figure*}

\clearpage

\begin{figure*}[ht!]
    \centering
	\includegraphics[width=\linewidth]{sup-main-pix3d-v2}
        {\footnotesize \hspace*{0.8cm} Input \hspace{1.9cm}  MGN  \hspace{1.6cm} LIEN \hspace{1.7cm}  InstPIFu \hspace{1.4cm}  Ours$_{\text{shape}}$ \hspace{1.0cm} Ours$_{\text{shape+texture}}$ \hspace{1.25cm} GT \hfill }
	\caption{
	\textbf{More qualitative results from \pixthreed~\cite{sun2018pix3d}.}
	}
	\label{fig:supp_qual_pix3d}
\end{figure*}

\clearpage

\end{document}